# Propositional Independence:
# Formula-Variable Independence and Forgetting

**Jérôme Lang**                                                              LANG@IRIT.FR
*IRIT-UPS, 118 route de Narbonne*
*31062 Toulouse Cedex, France*
**Paolo Liberatore**                                              LIBERATO@DIS.UNIROMA1.IT
*Dipartimento di Informatica e Sistemistica*
*Università di Roma "La Sapienza"*
*via Salaria 113, 00198 Roma, Italy*
**Pierre Marquis**                                       MARQUIS@CRIL.UNIV-ARTOIS.FR
*CRIL-CNRS/Université d'Artois*
*rue de l'Université*
*S.P. 16, 62307 Lens Cedex, France*

## Abstract

Independence – the study of what is relevant to a given problem of reasoning – has received an increasing attention from the AI community. In this paper, we consider two basic forms of independence, namely, a syntactic one and a semantic one. We show features and drawbacks of them. In particular, while the syntactic form of independence is computationally easy to check, there are cases in which things that intuitively are not relevant are not recognized as such. We also consider the problem of forgetting, i.e., distilling from a knowledge base only the part that is relevant to the set of queries constructed from a subset of the alphabet. While such process is computationally hard, it allows for a simplification of subsequent reasoning, and can thus be viewed as a form of compilation: once the relevant part of a knowledge base has been extracted, all reasoning tasks to be performed can be simplified.

## 1. Introduction

We successively present some motivations, the scope of the paper, and the contribution and the organization of the paper.

### 1.1 Motivations

In many situations of everyday life, we are confronted with the problem of *determining what is relevant*. Especially, as a preliminary step to various intelligent tasks (e.g., planning, decision making, reasoning), it is natural and reasonable to discard everything but what is relevant to achieve them efficiently. For instance, before starting to write this paper, we had to consider the relevant literature on the topic, and only it; gathering on our desks all the papers about relevance that we have at hand led us to set away our favourite cook books, because they are of no help to the task of writing this paper. The ability to focus on what is relevant (or dually to discard what is not) can be considered as a central, characteristic feature of intelligence; it is believed that over 90% of the neuronal connections of our brains are inhibitory and serve to give up sensorial inputs (Subramanian, Greiner, & Pearl, 1997). This explains why irrelevance, under various names as independence, irredundancy, influenceability, novelty, separability, is nowadays considered as an important notion in many fields of artificial intelligence (see a survey for more details, e.g., Greiner & Subramanian, 1995; Subramanian et al., 1997).

In the following, we are concerned with *relevance for reasoning*. In this framework, the task is typically that of determining whether some piece of knowledge (a query) $\varphi$ can be derived from a knowledge base $\Sigma$. Several reasoning schemes can be taken into account here, from the classical one (inference is classical entailment) to more sophisticated ones (common-sense inference). When dealing with such reasoning tasks, relevance is often exploited so as to make inference more efficient from a computational point of view. In other reasoning problems, the purpose is to explicitly derive some intensionally-characterized pieces of knowledge (e.g., tell me all what you know about Tweety). For such problems (that are not reducible to decision problems), relevance also has a role to play, for instance by allowing us to characterize the pieces of information we are interested in by forgetting the other ones.

To what extent is the goal of improving inference reachable? In order to address this point, a key issue is the *computational complexity* one. Indeed, assume that we know that the resolution of some reasoning problems can be sped up once relevant information has been elicited. In the situation where it is computationally harder to point out such information from the input than to reason directly from the input, computational benefits are hard to be expected. If so, alternative uses of relevance for reasoning are to be investigated. For instance, searching for relevance information can be limited by considering only pieces of knowledge that can be generated in a tractable way. If such information depend only on the knowledge base, another possible approach is to (tentatively) compensate the computational resources spent in deriving the relevance information through many queries (computing pieces of relevant information can then be viewed as a form of compilation).

## 1.2 Scope of the Paper

Little is known about the computational complexity of relevance. This paper contributes to fill this gap. The complexity of several logic-based relevance relations is identified in a propositional setting. By logic-based we mean that the notions of relevance we focus on are not extra-logical but built inside the logic: they are defined using the standard logical notions of (classical) formula, model, logical deduction, etc. The stress is laid on notions of relevance that can prove helpful for improving inference and, in particular, the most basic form of it, classical entailment. Relevance is captured by relations in the metalanguage of the logic, that is, we formalize relevance as a relation between objects of the propositional language (formulas or sets of literals/variables) thus expressing the fact that some object is relevant to some other one.

Two notions play a central role in this paper. The first one, *(semantical) formula-variable independence* (FV-independence for short) tells that a propositional formula $\Sigma$ is independent from a given set $V$ of variables if and only if it can be rewritten equivalently as a formula in which none of the variables in $V$ appears. The second one is the notion of *forgetting a given set $V$ of variables in a formula $\Sigma$*. It is intimately linked to the notion of formula-variable independence because, as we show, the result of forgetting the set of variables $V$ in a formula $\Sigma$ can be defined as the strongest consequence of $\Sigma$ being independent from $V$. Both notions have appeared in the literature under various names and with several different (but equivalent) definitions, and are highly useful for many tasks in automated reasoning and many problems in artificial intelligence:

1. *automated deduction and consequence finding.* Formula-variable independence can be useful for checking (un)satisfiability or for consequence finding. Indeed, structuring the knowledge base by finding useful independencies may be worth doing before running any search algorithm. This principle is at work in (Amir & McIlraith, 2000). In optimal cases, for example, a satisfiability problem will be decomposed into a small number of satisfiability problems on easier knowledge bases (with less variables) as shown by Park and Gelder (1996). As to improving inference, this can be particularly helpful in the situation where the set of queries under consideration is limited to formulas $\varphi$ that are (syntactically or semantically) independent from a set $V$ of variables.

2. *query answering* (see in particular Amir & McIlraith, 2000), *diagnosis* (see the work by Darwiche, 1998). Rendering a formula $\Sigma$ independent from a set $V$ of variables through variable forgetting gives rise to a formula that is query-equivalent to $\Sigma$ w.r.t. $V$ in the sense that every logical consequence $\varphi$ of $\Sigma$ that is independent from $V$ also is a logical consequence of $\Sigma$ once made independent from $V$, and the converse holds as well. Interestingly, the set of all the formulas independent from a set of variables is a stable production field (Siegel, 1987; Inoue, 1992), and focusing on such a production field is valuable for several reasoning schemes. For instance, in the consistency-based framework for diagnosis (Reiter, 1987), the queries we are interested in are the conflicts of the system to be diagnosed, i.e., the clauses that are independent from every variable used to represent the system, except the abnormality propositions used to encode the component failures.

3. *knowledge base structuring, topic-based reasoning.* Formula-variable independence is a key notion for decomposing a propositional knowledge base (KB for short), i.e., a finite set of propositional formulas, into smaller subbases. Such a decomposition is all the more valuable as the number of variables the subbases depends on is low. Optimally, a knowledge base $\Sigma = \{\varphi_1, ..., \varphi_n\}$ is fully decomposable if it can be written as $\Sigma = \Sigma_1 \cup \ldots \cup \Sigma_n$ where $\Sigma_i$ and $\Sigma_j$ depend on disjoint sets of variables for all $i \neq j$. Such decompositions were considered in several papers (Parikh, 1996; Amir & McIlraith, 2000; Marquis & Porquet, 2000) with somewhat different motivations. The most intuitive motivation for searching such decompositions is that it gives a better understanding of the knowledge base, by structuring it with respect to (possible disjoint but not necessarily) sets of topics (Marquis & Porquet, 2000).

4. *belief revision and inconsistency-tolerant reasoning.* Decomposing a propositional knowledge base $\Sigma$ into subbases $\{\Sigma_1, \ldots, \Sigma_n\}$ proves also to be relevant for defining inconsistency-tolerant relations as well as belief revision operators. The approach proposed by Chopra and Parikh (1999) proceeds as follows: the knowledge base $\Sigma$ is first partitioned into $\{\Sigma_1, \ldots, \Sigma_n\}$ such that the intersection of the languages of the subbases (i.e., the sets $DepVar(\Sigma_i)$ of variables the subbases depend on) are as small as possible; then $\varphi$ is inferred from $\Sigma$ w.r.t. the given partition if and only if the conjunction of all $\Sigma_i$ such that $DepVar(\Sigma_i) \cap DepVar(\varphi) \neq \emptyset$ is consistent and entails $\varphi$. In a revision situation, this approach also ensures that the only old beliefs that may be thrown away are about the variables relevant to the input formula. Finally, as recently shown by some of the authors of the present paper, forgetting can be advantageously exploited as a weakening mechanism for recovering consistency from an inconsistent KB (Lang & Marquis, 2002).

5. *belief update*. A belief update operator maps a knowledge base $\Sigma$ and an input formula $\varphi$ expressing some explicit evolution of the world to a new knowledge base $\Sigma \diamond \varphi$; $\Sigma$ and $\Sigma \diamond \varphi$ respectively represent the agent's knowledge *before* and *after* the evolution of the world expressed by the update. Several authors argued that an update operator should preserve the part of the knowledge base not concerned by the update. This leads to the following three-stage process, proposed independently by Doherty, Lukaszewicz, and Madalinska-Bugaj (1998) and by Herzig and Rifi (1998) (and named MPMA and WSS↓, respectively): (i) determine the variables relevant to the update, namely, $DepVar(\varphi)$; (ii) forget these variables in $\Sigma$ to obtain a new formula $ForgetVar(\Sigma, DepVar(\varphi))$; (iii) expand by $\varphi$. In more compact terms, this update operator is expressed by $\Sigma \diamond \varphi = ForgetVar(\Sigma, DepVar(\varphi)) \wedge \varphi$. Thus, both our results on FV-independence and variable forgetting are relevant to the computational issues pertaining to this kind of belief updates.

6. *reasoning about action, decision making, planning*. Logical languages for reasoning about action express the effects (deterministic or not, conditional or not) of actions by means of propositional formulas (Gelfond & Lifschitz, 1993; Sandewall, 1995; Fargier, Lang, & Marquis, 2000; Herzig, Lang, Marquis, & Polacsek, 2001); finding the variables the effects are dependent on enables to identify the variables whose truth value may be changed by the action; moreover, formula-literal independence – a refinement of FV-independence that we introduce – also tells us in which direction (from false to true and/or from true to false) the possible change may occur. This is especially useful for filtering out irrelevant actions in a decision making or planning problem.

7. *preference representation*. Logical languages for representing preference write elementary goals as propositional formulas and it is crucial to identify those variables that have no influence on the agent's preference. Therefore, formula-variable independence has an important role to play; for instance, the framework of so-called *ceteris paribus* preference statements of Tan and Pearl (1994) interprets a preference item $\varphi : \psi > \neg\psi$ by: for any pair of worlds $(\omega, \omega')$ such that (i) $\omega \models \psi$, (ii) $\omega' \models \neg\psi$ and (iii) $\omega$ and $\omega'$ coincide on all variables $\varphi$ *and* $\psi$ *are independent from*, then we have a strict preference of $\omega$ over $\omega'$.

## 1.3 Contribution and Organization of the Paper

The rest of the paper is structured as follows. After some formal preliminaries given in Section 2, the key notion of formula-variable independence is presented in Section 3. Because it captures a very basic form of logic-based independence, this relation has already been introduced in the literature under several names, like influenceability (Boutilier, 1994), relevance to a subject matter (Lakemeyer, 1997), or redundancy (Doherty et al., 1998). Although it is conceptually and technically simple, this notion has not been studied in a systematic way, and our very first contribution aims at filling this gap, first by giving several equivalent characterizations of this notion (this is useful, as several papers introduced and used the same concepts under different names), and second by investigating carefully its computational complexity. In particular, we show that, in the general case, checking whether a formula is independent from a set of variables is coNP-complete. Then, we go beyond this very simple notion by introducing the more fine-grained notion of *formula/literal independence* in order to discriminate the situation where a formula $\Sigma$ conveys some information about a literal but no information about its negation. This refinement is helpful whenever the polarity of information is significant, which is the case in many AI fields (including closed-world reasoning and reasoning about actions). We also study several interesting notions derived from formula-variable and formula-literal independence, such as the notion of *simplified formula* ($\Sigma$ is Lit- (Var-) simplified if it depends on every literal (variable) occurring in it) and the corresponding process of simplifying a formula. Despite this complexity and because the size of a simplified formula can never be larger than the size of the original formula, simplification can prove a valuable relevance-based preprocessing for improving many forms of inference.

In Section 4, we turn to the second key notion, namely *forgetting a given set of variables in a formula* (Lin & Reiter, 1994). The forgetting process plays an important role in many AI tasks and has been studied in the literature under various other names, such as variable elimination, or marginalization (Kohlas, Moral, & Haenni, 1999). Several semantical characterizations and metatheoretic properties of forgetting are presented. Based on this notion, we introduce an additional notion of dependence between two formulas given a set of variables: we state that $\Sigma$ is equivalent to $\Phi$ w.r.t. $V$ if and only if both formulas are logically equivalent once made independent from every variable except those of $V$. Here again, it is important to make a distinction between a variable and its negation (for instance, one can be interested in the positive conflicts of a system, only). For this purpose, we introduce a notion of literal forgetting that refines the corresponding notion of variable forgetting. We show how closed-world inference can be simply characterized from the corresponding equivalence relation. Finally, we identify the complexity of both notions of equivalence and show them to be hard ($\Pi_2^p$-complete). As a consequence, forgetting variables or literals within a formula cannot be achieved in polynomial time in the general case (unless the polynomial hierarchy collapses at the first level). We also show that a polysize propositional representation of forgetting is very unlikely to exist in the general case. We nevertheless present some restricted situations where forgetting is tractable.

Section 5 shows FV-independence closely related to notions of irrelevance already introduced in the literature by various authors. Section 6 discusses other related work and sketches further extensions of some notions and results studied in the paper. Finally, Section 7 concludes the paper. Proofs of the main propositions are reported in an appendix. A glossary of the notations is at the end of this paper, right before the bibliography.

## 2. Preliminaries

We first recall some basic notions from propositional logic, and from complexity theory.

### 2.1 Propositional Logic

Let $PS$ be a finite set of propositional variables. $PROP_{PS}$ is the propositional language built up from $PS$, the connectives and the Boolean constants $true$ and $false$ in the usual way. For every $V \subseteq PS$, $PROP_V$ denotes the sublanguage of $PROP_{PS}$ generated from the variables of $V$ only. A *literal* of $PROP_V$ is either a variable of $V$ (positive literal) or the negation of a variable of $V$ (negative literal). Their set is denoted $L_V$, while $L_V^+$ (resp. $L_V^-$) denotes the set of positive (resp. negative) literals built up from $V$. A clause $\delta$ (resp. a term $\gamma$) of $PROP_V$ is a (possibly empty) finite disjunction (resp. conjunction) of literals of $PROP_V$. A CNF (resp. a DNF) formula of $PROP_V$ is a finite conjunction of clauses (resp. disjunction of terms) of $PROP_V$. As usual, every finite set of formulas from $PROP_{PS}$ is identified with the formula that is the conjunction of its elements.

From now on, $\Sigma$ denotes a propositional formula, i.e., a member of $PROP_{PS}$. $Var(\Sigma)$ is the set of propositional variables appearing in $\Sigma$. If $L \subseteq L_{PS}$, $Var(L)$ is the set of variables from $PS$ upon which literals of $L$ are built. Elements of $PS$ are denoted $v$, $x$, $y$ etc. Elements of $L_{PS}$ are denoted $l$, $l_1$, $l_2$ etc. Subsets of $PS$ are denoted $V$, $X$, $Y$ etc. In order to simplify notations, we assimilate every singleton $V = \{v\}$ with its unique element $v$. The *size* of a formula $\Sigma$, denoted by $|\Sigma|$, is the number of occurrences of variables it contains. A propositional formula $\Sigma$ is said to be in Negation Normal Form (NNF) if and only if only propositional symbols are in the scope of an occurrence of $\neg$ in $\Sigma$. It is well-known that every propositional formula $\Sigma$ built up from the connectives $\wedge$, $\vee$, $\neg$, $\Rightarrow$ only, can be turned in linear time into an equivalent NNF formula by "pushing down" every occurrence of $\neg$ in it (i.e., exploiting De Morgan's law) and removing double negations (since $\neg$ is involutive). Slightly abusing words, we call the formula resulting from this normalization process *the NNF* of $\Sigma$ and we note $Lit(\Sigma)$ the set of literals from $L_{PS}$ occurring in the NNF of $\Sigma$. For instance, the NNF of $\Sigma = \neg((\neg a \wedge b) \vee c)$ is $(a \vee \neg b) \wedge \neg c$; hence, we have $Lit(\Sigma) = \{a, \neg b, \neg c\}$. Note that the NNF of a formula depends on its syntactical structure, i.e., two formulae that are synctatically different may have different NNFs even if they are equivalent.

Full instantiations of variables of $V \subseteq PS$ are called $V$-*worlds*[1]; they are denoted by $\omega_V$ and their set is denoted $\Omega_V$. When $A$ and $B$ are two disjoint subsets of $PS$, $\omega_A \wedge \omega_B$ denotes the $A \cup B$-world that coincides with $\omega_A$ on $A$ and with $\omega_B$ on $B$. An interpretation $\omega$ over $PROP_{PS}$ is just a $PS$-world, and $\omega$ is said to be a model of $\Sigma$ whenever it makes $\Sigma$ true. We denote $Mod(\Sigma)$ the set of models of $\Sigma$.

For every formula $\Sigma$ and every variable $x$, $\Sigma_{x \leftarrow 0}$ (resp. $\Sigma_{x \leftarrow 1}$) is the formula obtained by replacing every occurrence of $x$ in $\Sigma$ by the constant $false$ (resp. $true$). $\Sigma_{l \leftarrow 1}$ (resp. $\Sigma_{l \leftarrow 0}$) is an abbreviation for $\Sigma_{x \leftarrow 1}$ (resp. $\Sigma_{x \leftarrow 0}$) when $l$ is a positive literal $x$ and for $\Sigma_{x \leftarrow 0}$ (resp. $\Sigma_{x \leftarrow 1}$) when $l$ is a negative literal $\neg x$.

Given an interpretation $\omega$ and a literal $l$, we let $Force(\omega, l)$ denote the interpretation that gives the same truth value as $\omega$ to all variables except the variable of $l$, and such that $Force(\omega, l) \models l$. In other words, $Force(\omega, l)$ is the interpretation satisfying $l$ that is the closest to $\omega$. For instance, provided that $PS = \{a, b\}$ and $\omega(a) = \omega(b) = 1$, we have $Force(\omega, \neg b)(a) = 1$ and $Force(\omega, \neg b)(b) = 0$. Clearly, if $\omega \models l$ then $Force(\omega, l) = \omega$. If $L = \{l_1, \ldots, l_n\}$ is a consistent set of literals, then $Force(\omega, L)$ is defined as

$$Force(\ldots(Force(\omega, l_1), \ldots), l_n).$$

---

1. The $V$-worlds are also called partial models over $V$.

Lastly, given an interpretation $\omega$ and a variable $x$, we let $Switch(\omega, x)$ denote the interpretation that gives the same truth value as $\omega$ to all variables except $x$, and that gives to $x$ the value opposite to that given by $\omega$.

Two formulas $\Psi$ and $\Phi$ are said to be *equivalent modulo* a formula $\Sigma$ if and only if $\Sigma \wedge \Psi \equiv \Sigma \wedge \Phi$.

In this paper we use the concepts of prime implicates and prime implicants. The set of prime implicates of a formula $\Sigma$, denoted by $IP(\Sigma)$, is defined as:

$$IP(\Sigma) = \{\delta \text{ clause} \mid \Sigma \models \delta \text{ and } \not\exists \delta' \text{ clause s.t. } \Sigma \models \delta' \text{ and } \delta' \models \delta \text{ and } \delta \not\models \delta'\}.$$

Among all the implicates of $\Sigma$ (i.e., the clauses entailed by $\Sigma$), the prime implicates of $\Sigma$ are the minimal ones w.r.t. $\models$ (i.e., the logically strongest ones). The set of prime implicants of a formula $\Sigma$, denoted by $PI(\Sigma)$, is defined dually as:

$$PI(\Sigma) = \{\gamma \text{ term} \mid \gamma \models \Sigma \text{ and } \not\exists \gamma' \text{ term s.t. } \gamma' \models \Sigma \text{ and } \gamma \models \gamma' \text{ and } \gamma' \not\models \gamma\}.$$

Among all the implicants of $\Sigma$ (i.e., the terms implying $\Sigma$), the prime implicants of $\Sigma$ are the maximal ones w.r.t. $\models$ (i.e., the logically weakest ones).

Of course, the set of prime implicants/ates may contain equivalent terms/clauses. We can restrict our attention to one term/clause for each set of equivalent terms/clauses. Stated otherwise, in $PI(\Sigma)$ and $IP(\Sigma)$, only one representative per equivalence class is kept.

**Example 1** *Let $\Sigma = \{a \vee b, \neg a \wedge c \Rightarrow e, d \Leftrightarrow e\}$. The set of prime implicates of $\Sigma$ is, by definition,*

$$IP(\Sigma) = \{a \vee b, a \vee \neg c \vee e, \neg d \vee e, d \vee \neg e, a \vee \neg c \vee d\}.$$

## 2.2 Computational Complexity

The complexity results we give in this paper refer to some complexity classes which deserve some recalls. More details can be found in Papadimitriou's (1994) textbook. Given a problem A, we denote by $\overline{A}$ the complementary problem of A. We assume that the classes P, NP and coNP are known to the reader. The following classes will also be considered:

- BH$_2$ (also known as DP) is the class of all languages $L$ such that $L = L_1 \cap L_2$, where $L_1$ is in NP and $L_2$ in coNP. The canonical BH$_2$-complete problem is SAT–UNSAT: a pair of formulas $\langle \varphi, \psi \rangle$ is in SAT–UNSAT if and only if $\varphi$ is satisfiable and $\psi$ is not. The complementary class coBH$_2$ is the class of all languages $L$ such that $L = L_1 \cup L_2$, where $L_1$ is in NP and $L_2$ in coNP. The canonical coBH$_2$-complete problem is SAT-OR-UNSAT: a pair of formulas $\langle \varphi, \psi \rangle$ is in SAT-OR-UNSAT if and only if $\varphi$ is satisfiable or $\psi$ is not.

- $\Delta_2^p = \mathsf{P}^{\mathsf{NP}}$ is the class of all languages recognizable in polynomial time by a deterministic Turing machine equipped with an $\mathsf{NP}$ oracle, i.e., a device able to solve any instance of an $\mathsf{NP}$ or a $\mathsf{coNP}$ problem in unit time. $\mathsf{F}\Delta_2^p$ is the corresponding class of function problems.

- $\Sigma_2^p = \mathsf{NP}^{\mathsf{NP}}$ is the class of all languages recognizable in polynomial time by a nondeterministic Turing machine equipped with an $\mathsf{NP}$ oracle. The canonical $\Sigma_2^p$-complete problem 2-QBF is the set of all triples $\langle A = \{a_1, ..., a_m\}, B = \{b_1, ..., b_n\}, \Phi \rangle$ where $A$ and $B$ are two disjoint sets of propositional variables and $\Phi$ is a formula from $PROP_{A \cup B}$. A positive instance of this problem is a triple $\langle A, B, \Phi \rangle$ for which there exists a $A$-world $\omega_A$ such that for all $B$-world $\omega_B$ we have $\omega_A \wedge \omega_B \models \Phi$.

- $\Pi_2^p = \mathsf{co}\Sigma_2^p = \mathsf{coNP}^{\mathsf{NP}}$. The canonical $\Pi_2^p$-complete problem 2-$\overline{\text{QBF}}$ is the set of all triples $\langle A = \{a_1, ..., a_m\}, B = \{b_1, ..., b_n\}, \Phi \rangle$ where $A$ and $B$ are two disjoint sets of propositional variables and $\Phi$ is a formula from $PROP_{A \cup B}$. A positive instance of this problem is a triple $\langle A, B, \Phi \rangle$ such that for every $A$-world $\omega_A$ there exists a $B$-world $\omega_B$ for which $\omega_A \wedge \omega_B \models \Phi$.

$\Sigma_2^p$ and $\Pi_2^p$ are complexity classes located at the so-called second level of the polynomial hierarchy, which plays a prominent role in knowledge representation and reasoning.

## 3. Formula-Literal and Formula-Variable Independence

Formula-literal and formula-variable independence capture some forms of independence between the truth values of variables and the possible truth values of a formula. Roughly speaking, $\Sigma$ is dependent on $l$ because it tells one something positive about $l$: more precisely, there is a contest, i.e., a conjunction of literals, that, added to to $\Sigma$ enables one to infer $l$. For instance, $\Sigma = (a \Rightarrow b)$ is dependent on $b$, since $b$ can be inferred from $\Sigma$ under the assumption that $a$ is true. Of course, we cannot assume contexts that are inconsistent with $\Sigma$, and we cannot assume $l$ itself. A formula $\Sigma$ will be considered as independent from variable $x$ if and only if it is both independent from $x$ and independent from $\neg x$. Dually, we can interpret both forms of dependence as aboutness relations; when $\Sigma$ is dependent on a literal $l$, it tells one something about $l$, and when $\Sigma$ is dependent on a variable $x$, it tells something about $x$ or about $\neg x$.

### 3.1 Syntactical Independence

The easiest way to define dependence between a formula $\Sigma$ and a literal $l$ is by assuming that $\Sigma$, when put into NNF, contains $l$. Reminding that $Lit(\Sigma)$ is the set of literals that occur in the NNF of $\Sigma$, this can be formally expressed by the following definition:

**Definition 1 (syntactical FL-independence)** *Let $\Sigma$ be a formula from $PROP_{PS}$, $l$ a literal of $L_{PS}$, and $L$ a subset of $L_{PS}$.*

- *$\Sigma$ is said to be* syntactically Lit-dependent *on $l$ (resp. syntactically Lit-independent from $l$) if and only if $l \in Lit(\Sigma)$ (resp. $l \notin Lit(\Sigma)$).*

- $\Sigma$ *is said to be* syntactically Lit-dependent *on $L$ if and only if there is a $l \in L$ such that $\Sigma$ is syntactically Lit-dependent on $l$. Otherwise, $\Sigma$ is said to be* syntactically Lit-independent *from $L$.*

From this definition it follows immediately that $\Sigma$ is syntactically Lit-independent from $L$ if and only if $Lit(\Sigma) \cap L = \emptyset$. Thus, in order to determine whether a formula $\Sigma$ is syntactically Lit-dependent on a set $L$ of literals, it suffices to check whether there exists a literal in $L$ that occurs in the NNF of $\Sigma$.

**Example 2** *Let $\Phi = \neg(a \wedge b)$. The NNF of $\Phi$ is $(\neg a \vee \neg b)$; therefore, $\Phi$ is syntactically Lit-dependent on both $\neg a$ and $\neg b$, while it is syntactically Lit-independent from $a$ and $b$.*

As this example illustrates, a propositional formula can easily be syntactically Lit-independent from a literal while syntactically Lit-dependent on its negation. This is as expected, since $\Sigma$ may imply $l$ in some context, while it may be always impossible to derive $\neg l$. In other words, the set of literals $\Sigma$ is syntactically Lit-independent from is not closed under negation. Interestingly, a notion of syntactical formula-variable independence can be defined from the more basic notion of syntactical FL-independence.

**Definition 2 (syntactical FV-independence)** *Let $\Sigma$ be a formula from $PROP_{PS}$, $v$ a variable of $PS$, and $V$ a subset of $PS$.*

- $\Sigma$ *is said to be* syntactically Var-dependent *on $v$ (resp.* syntactically Var-independent *from $v$) if and only if $v \in Var(\Sigma)$ (resp. $v \notin Var(\Sigma)$).*

- $\Sigma$ *is said to be* syntactically Var-dependent *on $V$ if and only if there is a variable $v$ in $V$ s.t. $\Sigma$ is Var-dependent on it, i.e., if and only if $Var(\Sigma) \cap V \neq \emptyset$. Otherwise, $\Sigma$ is said to be* syntactically Var-independent *from $V$.*

**Example 3** *Let $\Sigma = (a \wedge \neg b)$. $\Sigma$ is syntactically Var-dependent on $a$ and on $b$ and syntactically Var-independent from $c$.*

Syntactical FV-independence can be easily expressed as syntactical FL-independence: $\Sigma$ is syntactically Var-independent from $V$ if and only if it is syntactically Lit-independent from $V \cup \{\neg x \mid x \in V\}$. Clearly enough, both syntactical formula-literal independence and syntactical formula-variable independence can be checked in linear time.

However, these basic forms of independence suffer from two important drawbacks. First, they do not satisfy the principle of irrelevance of syntax: two equivalent formulas are not always syntactically independent from the same literals/variables. Second, syntactical dependence does not always capture the intuitive meaning of dependence: for instance, $\Sigma = (a \wedge \neg b \wedge (a \vee b))$ is syntactically Lit-dependent on $a$, $\neg b$, $b$; since $\neg b$ can be derived from $\Sigma$, $\Sigma$ is about $\neg b$ in some sense. Contrastingly, there is no way to derive $b$ from $\Sigma$, unless producing an inconsistency.

Handling such a separation requires a more robust notion of independence, to be introduced in the following section.

## 3.2 Semantical Independence

We now give a *semantical* definition of independence, which does not suffer from the afore-mentioned drawbacks, i.e., a definition that does not depend on the syntactical form in which formulas are expressed. We will prove that this semantical definition of independence does not suffer from the second drawback of syntax independence, i.e., a formula that is semantically dependent on a literal always enables one to derive the literal in some context.

**Definition 3 ((semantical) FL-independence)** *Let $\Sigma$ be a formula from $PROP_{PS}$, $l \in L_{PS}$, and $L$ a subset of $L_{PS}$.*

- *$\Sigma$ is said to be Lit-independent[2] from $l$, denoted $l \not\mapsto \Sigma$, if and only if there exists a formula $\Phi$ s.t. $\Phi \equiv \Sigma$ and $\Phi$ is syntactically Lit-independent from $l$. Otherwise, $\Sigma$ is said to be Lit-dependent on $l$, denoted $l \mapsto \Sigma$. The set of all literals of $L_{PS}$ such that $l \mapsto \Sigma$ is denoted by $DepLit(\Sigma)$.*

- *$\Sigma$ is said to be Lit-independent from $L$, denoted $L \not\mapsto \Sigma$, if and only if $L \cap DepLit(\Sigma) = \emptyset$. Otherwise, $\Sigma$ is said to be Lit-dependent on $L$, denoted $L \mapsto \Sigma$.*

Simply rewriting the definition, $\Sigma$ is Lit-independent from $L$ if and only if there exists a formula $\Phi$ s.t. $\Phi \equiv \Sigma$ and $\Phi$ is syntactically Lit-independent from $L$. Thus, FL-independence is not affected by the syntactic form in which a formula is expressed, that is, replacing $\Sigma$ with any of its equivalent formulas does not modify the relation $\mapsto$. Since $\Sigma$ is Lit-independent from $L$ if and only if $\Sigma$ can be made syntactically Lit-independent from $L$ while preserving logical equivalence, it follows that syntactical Lit-independence implies Lit-independence, but the converse does not hold in the general case.

**Example 4** *Let $\Sigma = (a \wedge \neg b \wedge (a \vee b))$. We have $DepLit(\Sigma) = \{a, \neg b\}$. Note that $\Sigma$ is Lit-independent from $b$ because it is equivalent to $\Phi = (a \wedge \neg b)$, in which $b$ does not appear positively.*

As in the case of syntactical independence, we can formalize the fact that a formula $\Sigma$ has some effects on the truth value of a variable $v$. Indeed, we define a notion of (semantical) formula-variable independence, which can also be easily defined from the (semantical) notion of FL-independence.

**Definition 4 ((semantical) FV independence)** *Let $\Sigma$ be a formula from $PROP_{PS}$, $v \in PS$, and $V$ a subset of $PS$.*

- *$\Sigma$ is said to be Var-independent from $v$, denoted $v \not\mapsto^+_- \Sigma$, if and only if there exists a formula $\Phi$ s.t. $\Phi \equiv \Sigma$ and $\Phi$ is syntactically Var-independent from $v$. Otherwise, $\Sigma$ is said to be Var-dependent on $v$, denoted $v \mapsto^+_- \Sigma$. We denote by $DepVar(\Sigma)$ the set of all variables $v$ such that $v \mapsto^+_- \Sigma$.*

---

2. In order to avoid any ambiguity, we will refer to the syntactical forms of independence explicitly from now on; in other words, independence must be taken as semantical by default in the rest of the paper.

- $\Sigma$ is said to be Var-independent *from $V$, denoted $V \not\mapsto_-^+ \Sigma$, if and only if $V \cap DepVar(\Sigma) = \emptyset$. Otherwise, $\Sigma$ is said to be* Var-dependent *on $V$, denoted $V \mapsto_-^+ \Sigma$.*

Clearly, $\Sigma$ is Var-independent from $V$ if and only if there exists a formula $\Phi$ s.t. $\Phi \equiv \Sigma$ and $\Phi$ is syntactically Var-independent from $V$. Moreover, Var-independence is to Lit-independence as syntactical Var-independence is to syntactical Lit-independence; indeed, $\Sigma$ is Var-independent from $V$ if and only if $\Sigma$ is Lit-independent from $L_V$.

**Example 5** *Let $\Sigma = (a \wedge \neg b \wedge (a \vee c))$. We have $DepVar(\Sigma) = \{a, b\}$. Note that $\Sigma$ is Var-independent from $c$.*

The definition of semantical FL-independence is based on the set of literals of formulas equivalent to $\Sigma$. Intuitively, this is the easiest way to define a notion of independence that is not dependent on the syntax. However, proving theorems directly from this definition is not so easy. For instance, we will prove that determining whether a formula $\Sigma$ is Lit-dependent on literal $l$ is in NP, but this result cannot be *directly* proved from Definition 3, since checking all possible formulas equivalent to $\Sigma$ cannot be done with a polynomial non-deterministic guessing. We give now a semantical characterization of FL-independence.

**Proposition 1** *A formula $\Sigma$ is Lit-independent from literal $l$ if and only if, for any interpretation $\omega \in \Omega_{PS}$, if $\omega \models \Sigma$ then $Force(\omega, \neg l) \models \Sigma$.*

As a direct corollary, we get that $\Sigma$ is Lit-independent from $L$ if and only if for any literal $l \in L$ and any interpretation $\omega \in \Omega_{PS}$, if $\omega \models \Sigma$ then $Force(\omega, \neg l) \models \Sigma$.

This property gives an idea of how FL-dependence works. Indeed, if $\Sigma$ is Lit-dependent on a literal $l$, then there exists an interpretation $\omega$ such that $\omega \models \Sigma$ and $Force(\omega, \neg l) \not\models \Sigma$, which means that (a) $\omega \models l$ and (b) the literal $l$ in $\omega$ is "really needed" to make $\omega$ a model of $\Sigma$, that is, the partial interpretation obtained by removing $l$ in $\omega$ does not satisfy $\Sigma$.

This property also explains why FL-dependence formalizes the concept of "true in some context", as explained at the beginning of this section. Indeed, $\Sigma$ is Lit-dependent on $l$ if and only if there is some context (consistent with $\Sigma$, and that does not imply $l$), in which $\Sigma$ implies $l$. This can be proved from the above proposition: if $\omega$ is an interpretation such that $\omega \models \Sigma$ but $Force(\omega, \neg l) \not\models \Sigma$, then the term:

$$\gamma = \bigwedge(\{x \in PS \mid \omega \models \Sigma \wedge x\} \cup \{\neg x \mid x \in PS \text{ and } \omega \models \Sigma \wedge \neg x\}) \setminus \{l\}$$

is consistent with $\Sigma$ (by construction, $\gamma$ is equivalent to the disjunction of a term equivalent to $\omega$ with a term equivalent to $Force(\omega, \neg l)$); it also holds that $\gamma \wedge \Sigma \models l$, while $\gamma \not\models l$, that is, $\gamma$ is a context in which $\Sigma$ implies $l$.

Moreover, Proposition 1 shows that our notion of Lit-independence coincides with the notion of *(anti-)monotonicity* (Ryan, 1991, 1992). To be more precise, a (consistent) formula is said to be monotonic (resp. antimonotonic) in variable $v$ if and only if it is Lit-independent from $\neg v$ (resp. from $v$). Interestingly, a subclassical[3] inference relation, called *natural inference*, has been defined on this ground (Ryan, 1991, 1992). Basically, a formula $\varphi$ is considered as a consequence of a formula $\Sigma$ if and only if it is a logical consequence of it, and $\Sigma$ Lit-independence of a literal implies its $\varphi$ Lit-independence. Accordingly, natural inference prevents us from considering $p \lor q$ as a consequence of $p$ (it is a relevant implication). All our characterization results about Lit-independence, including complexity results, have an immediate impact on such a natural inference relation (especially, they directly show that the complexity of natural inference is in $\Delta_2^p$).

---

3. That is, a restriction of classical entailment $\models$.

Proposition 1 is easily extended to formula-variable independence:

**Corollary 1** *A formula $\Sigma$ is Var-independent from variable $v$ if and only if, for any interpretation $\omega \in \Omega_{PS}$, we have $\omega \models \Sigma$ if and only if $Switch(\omega, v) \models \Sigma$.*

The following metatheoretic properties of FL-independence are used in the rest of this section.

**Proposition 2**
*(1) $DepLit(\Sigma) \subseteq Lit(\Sigma)$;*
*(2) If $\Sigma \equiv \Phi$, then $DepLit(\Sigma) = DepLit(\Phi)$;*
*(3a) $DepLit(\Sigma \land \Phi) \subseteq DepLit(\Sigma) \cup DepLit(\Phi)$;*
*(3b) $DepLit(\Sigma \lor \Phi) \subseteq DepLit(\Sigma) \cup DepLit(\Phi)$;*
*(4) $l \in DepLit(\Sigma)$ if and only if $\neg l \in DepLit(\neg\Sigma)$.*

FV-independence exhibits similar properties, plus negation stability (point (4) below), which is not satisfied by FL-independence.

**Proposition 3**
*(1) $DepVar(\Sigma) \subseteq Var(\Sigma)$;*
*(2) If $\Sigma \equiv \Phi$, then $DepVar(\Sigma) = DepVar(\Phi)$;*
*(3a) $DepVar(\Sigma \land \Phi) \subseteq DepLit(\Sigma) \cup DepLit(\Phi)$;*
*(3b) $DepVar(\Sigma \lor \Phi) \subseteq DepLit(\Sigma) \cup DepLit(\Phi)$;*
*(4) $DepVar(\neg\Sigma) = DepVar(\Sigma)$.*

Beyond these properties, FL-independence and FV-independence do not exhibit any particularly interesting structure. In particular, FL-independence and FV-independence neither are monotonic nor anti-monotonic w.r.t. expansion of $\Sigma$ (strengthening or weakening $\Sigma$ can easily make it no longer independent from a set of literals or variables).

Recall that $L \mapsto \Sigma$ if and only if $L \cap DepLit(\Sigma) \neq \emptyset$. This means that the Lit-dependence of $\Sigma$ on $L$ only implies the Lit-dependence on a literal of $L$, not a "full" Lit-dependence on any literal of $L$. In other words, if we want to check whether a formula $\Sigma$ is Lit-dependent on *any* literal of $L$, we need a notion stronger than FL-dependence, called full FL-dependence. The same can be said for FV-dependence.

**Definition 5 (full FL/FV-dependence)** *Let $\Sigma$ be a formula from $PROP_{PS}$, $L$ be a subset of $L_{PS}$ and $V$ be a subset of $PS$.*

- *$\Sigma$ is fully Lit-dependent on $L$ if and only if $L \subseteq DepLit(\Sigma)$.*

- *$\Sigma$ is fully Var-dependent on $V$ if and only if $V \subseteq DepVar(\Sigma)$.*

**Example 6** $\Sigma = (a \wedge \neg b \wedge (b \vee \neg b))$ *is fully Lit-dependent on* $\{a, \neg b\}$ *and fully Var-dependent on* $\{a, b\}$*. Contrastingly,* $\Sigma$ *is not fully Lit-dependent on* $\{a, b\}$*.*

Clearly enough, whenever $\Sigma$ is fully Lit-dependent on $L$, it is also fully Var-dependent on $Var(L)$. However, the converse does not hold, as the previous example shows.

While full FL-dependence (resp. full FV-dependence) can be checked in linear time once $DepLit(\Sigma)$ (resp. $DepVar(\Sigma)$) is known, at the end of the section we prove that determining these sets is NP-hard, and that deciding full FL-dependence (as well as full FV-dependence) is NP-complete.

Let us now consider the particular case of full FL-dependence when $L = Lit(\Sigma)$, or the full FV-dependence when $V = Var(\Sigma)$. If full dependence holds in these cases, we say that $\Sigma$ is Lit-simplified, or Var-simplified, respectively. Var-simplification is achieved when $\Sigma$ contains no occurrence of any variable it is Var-independent from. Lit-simplification corresponds to the more restricted situation where the NNF of $\Sigma$ does not contain any occurrence of a literal it is Lit-independent from.

If a formula $\Sigma$ is not Lit-simplified (resp. Var-simplified), then there is some literal (resp. variable) that occurs in the NNF of $\Sigma$, but $\Sigma$ is not Lit-dependent (resp. Var-dependent) on it. This means that the syntactic form in which $\Sigma$ is expressed contains a literal or variable that is indeed useless.

**Definition 6 (simplified formula)** *Let* $\Sigma$ *be a formula from* $PROP_{PS}$*.*

- $\Sigma$ *is* Lit-simplified *if and only if* $Lit(\Sigma) = DepLit(\Sigma)$*.*

- $\Sigma$ *is* Var-simplified *if and only if* $Var(\Sigma) = DepVar(\Sigma)$*.*

As the following example illustrates, every formula that is Lit-simplified also is Var-simplified but the converse does not hold in the general case.

**Example 7** $\Sigma = (a \wedge \neg b \wedge (b \vee \neg b) \wedge (a \vee c))$ *neither is Lit-simplified nor Var-simplified. The equivalent formula* $(a \wedge \neg b \wedge (b \vee \neg b))$ *is Var-simplified but it is not Lit-simplified. Finally, the equivalent formula* $(a \wedge \neg b)$ *is both Lit-simplified and Var-simplified.*

Simplified formulas do not incorporate any useless literals or variables. As the next proposition shows it, the notion of simplified formula actually is the point where syntactical independence and (semantical) independence coincide.

**Proposition 4** *Let* $\Sigma$ *be a formula from* $PROP_{PS}$*.*

- $\Sigma$ *is* Lit-simplified *if and only if the following equivalence holds: for every* $L \subseteq L_{PS}$*,* $\Sigma$ *is syntactically Lit-independent from* $L$ *if and only if* $L \not\rightarrow \Sigma$ *holds.*

- $\Sigma$ *is* Var-simplified *if and only if the following equivalence holds: for every* $V \subseteq PS$*,* $\Sigma$ *is syntactically Var-independent from* $V$ *if and only if* $V \not\rightarrow_{-}^{+} \Sigma$ *holds.*

Thus, while a formula can easily be Lit-independent from a set of literals without being syntactically Lit-independent from it, simplification is a way to join Lit-independence with its syntactical restriction (which is easier to grasp, and as we will see soon, easier to check in the general case): Lit-independence and syntactical Lit-independence coincide on Lit-simplified formulas. The same holds for Var-independence and syntactical Var-independence.

The strength of the notion of simplification lies in the fact that every formula can be simplified preserving its models. This is useful, as simplified formulas can be shorter and easier to understand than formulas containing useless literals.

**Proposition 5** *For every* $\Sigma$ *from* $PROP_{PS}$, *there exists a Lit-simplified formula* $\Phi$ *s.t.* $\Sigma \equiv \Phi$.

Since Lit-simplified KBs are also Var-simplified, this proposition also shows that every KB can be Var-simplified without modifying its set of models.

Interestingly, both FL-independence and FV-independence can be characterized without considering the corresponding syntactic notions of independence, just by comparing formulas obtained by setting the truth value of literals we want to check the dependence.

**Proposition 6** *Let* $\Sigma$ *be a formula from* $PROP_{PS}$ *and* $l$ *be a literal of* $L_{PS}$. *The next four statements are equivalent:*
*(1)* $l \not\to \Sigma$;
*(2)* $\Sigma_{l\leftarrow 1} \models \Sigma_{l\leftarrow 0}$;
*(3)* $\Sigma \models \Sigma_{l\leftarrow 0}$;
*(4)* $\Sigma_{l\leftarrow 1} \models \Sigma$.

The above properties can be used to check whether a formula is Lit-dependent on a literal, as the following example shows.

**Example 8** $\Sigma = (a \wedge \neg b \wedge (b \vee \neg b))$ *is Lit-independent from* $b$ *since replacing* $b$ *by true within* $\Sigma$ *gives rise to an inconsistent formula. Contrastingly,* $\Sigma$ *is Lit-dependent on* $\neg b$ *since replacing* $b$ *by false within* $\Sigma$ *gives* $\Sigma_{b\leftarrow 0} \equiv a$, *which is not at least as logically strong as the inconsistent formula obtained by replacing* $b$ *by true within* $\Sigma$.

A similar proposition holds for FV-independence, characterizing the variables a formula $\Sigma$ depends on, using the formulas $\Sigma_{x\leftarrow 0}$ and $\Sigma_{x\leftarrow 1}$.

**Proposition 7** *Let* $\Sigma$ *be a formula from* $PROP_{PS}$ *and* $x$ *be a variable of* $PS$. *The next four statements are equivalent:*
*(1)* $x \not\to^+_- \Sigma$;
*(2)* $\Sigma_{x\leftarrow 0} \equiv \Sigma_{x\leftarrow 1}$;
*(3)* $\Sigma \equiv \Sigma_{x\leftarrow 0}$;
*(4)* $\Sigma \equiv \Sigma_{x\leftarrow 1}$.

As in the case of literal dependence, the above property can be used to find out whether a formula is Var-dependent on a variable.

**Example 9** $\Sigma = (a \wedge (b \vee \neg b))$ *is Var-independent from* $b$ *since we have* $\Sigma_{b\leftarrow 0} \equiv \Sigma_{b\leftarrow 1} \equiv a$.

Interestingly, FL-independence and FV-independence can be determined in an efficient way when $\Sigma$ is given in some specific normal forms, namely, prime implicate normal form or prime implicant normal form. For such normal forms, Lit-independence and Var-independence come down to their corresponding syntactical forms.

**Proposition 8** *Let $\Sigma$ be a formula from $PROP_{PS}$ and $L$ be a subset of $L_{PS}$. The next statements are equivalent:*

*(1) $L \not\hookrightarrow \Sigma$;*
*(2) $PI(\Sigma) \subseteq \{\gamma \mid \gamma$ is a term that does not contain any literal from $L\}$;*
*(3) $IP(\Sigma) \subseteq \{\delta \mid \delta$ is a clause that does not contain any literal from $L\}$.*

**Proposition 9** *Let $\Sigma$ be a formula from $PROP_{PS}$ and $V$ be a subset of $PS$. The next statements are equivalent:*

*(1) $V \not\hookrightarrow^+ \Sigma$;*
*(2) $PI(\Sigma) \subseteq \{\gamma \mid \gamma$ is a term that does not contain any variable from $V\}$;*
*(3) $IP(\Sigma) \subseteq \{\delta \mid \delta$ is a clause that does not contain any variable from $V\}$.*

**Example 10** *Let $\Sigma = (a \wedge \neg b \wedge (b \vee \neg b))$. We have $PI(\Sigma) = \{(a \wedge \neg b)\}$ and $IP(\Sigma) = \{a, \neg b\}$ (up to logical equivalence). We can easily observe that $\Sigma$ is Lit-independent from $b$ looking at $PI(\Sigma)$ (or $IP(\Sigma)$). We can also easily state that $\Sigma$ is Lit-dependent on $\neg b$ and Var-independent from $c$ by considering any of these normal forms.*

Proposition 7 shows that a KB can be easily Var-simplified (i.e., in polynomial time) as soon as the variables it is Var-independent from have been determined. Indeed, we can easily design a greedy algorithm for simplifying KBs. This algorithm consists in considering every variable $x$ of $Var(\Sigma)$ in a successive way, while replacing $\Sigma$ by $\Sigma_{x \leftarrow 0}$ whenever $\Sigma$ is Var-independent from $x$. This algorithm runs in time polynomial in the size of $\Sigma$ once the variables $\Sigma$ is Var-independent from have been computed. At the end, the resulting KB is Var-simplified.

Lit-simplifying a KB is not so easy if no assumptions are made about its syntax, due to the fact that literals signs are crucial here. Indeed, looking only at the occurrence of a variable inside a formula is not sufficient to state whether or not this is a positive occurrence (or a negative one). Fortunately, turning a formula into its NNF is computationally easy (as long as it does not contain any occurrence of connectives like $\Leftrightarrow$ or $\oplus$) and it proves sufficient to design a greedy algorithm for Lit-simplifying a KB when the literals it is Lit-independent from have been identified. Indeed, within an NNF formula, every literal can be considered easily as an atomic object. This algorithm consists in considering every literal $l$ of $Lit(\Sigma)$ in a successive way, while replacing every occurrence of $l$ in $\Sigma$ by $false$ considering every literal of $Lit(\Sigma)$ as an atom. Stated otherwise, when $l$ is a positive (resp. negative) literal $x$ (resp. $\neg x$), replacing $l$ by $false$ does not mean replacing $\neg x$ (resp. $x$) by $true$: only the occurrences of $l$ with the right sign are considered. This algorithm runs in time polynomial in the size of $\Sigma$ once the literals $\Sigma$ is Lit-independent from have been computed. At the end, the resulting KB is Lit-simplified.

## 3.3 Complexity Results

While syntactical FL and FV-dependence can be easily checked in linear time in the size of the input, this is far from being expected for (semantical) FL-dependence and FV-dependence in the general case:

**Proposition 10 (complexity of FL/FV-dependence)** FL DEPENDENCE, FV DEPENDENCE, FULL FL DEPENDENCE *and* FULL FV DEPENDENCE *are* NP-*complete*.

Thus, although they look simple, the problems of determining whether a formula is independent from a literal or a variable are as hard as checking propositional entailment. Interestingly, the complexity of both decision problems fall down to P whenever checking (in)dependence becomes tractable. Apart from the case of syntactical independence, some other restrictions on $\Sigma$ makes (in)dependence testable in polynomial time. Especially, we get:

**Proposition 11** *Whenever $\Sigma$ belongs to a class $\mathcal{C}$ of CNF formulas that is tractable for clausal query answering (i.e., there exists a polytime algorithm to determine whether $\Sigma \models \gamma$ for any CNF formula $\gamma$) and stable for variable instantiation (i.e., replacing in $\Sigma \in \mathcal{C}$ any variable by* true *or by* false *gives a formula that still belongs to $\mathcal{C}$) then* FL DEPENDENCE, FV DEPENDENCE, FULL FL DEPENDENCE *and* FULL FV DEPENDENCE *are in* P.

In particular, when $\Sigma$ is restricted to a renamable Horn CNF formula or to binary clauses (Krom formula), all four decision problems above belong to P.

We have also investigated the complexity of checking whether a formula is Lit-simplified (and Var-simplified):

**Proposition 12** LIT-SIMPLIFIED FORMULA *and* VAR-SIMPLIFIED FORMULA *are* NP-*complete*.

All these complexity results have some impact on the approaches that explicitly need computing $DepVar(\Sigma)$ as a preprocessing task. Namely, we have the following result:

**Proposition 13**

1. *Determining whether $DepLit(\Sigma) = L$ (where $L$ is a set of literals), and determining whether $DepVar(\Sigma) = X$ (where $X$ is a set of variables) is* BH$_2$-*complete*.

2. *The search problem consisting in computing $DepLit(\Sigma)$ (respectively $DepVar(\Sigma)$) is in* F$\Delta_2^p$ *and is both* NP-*hard and* coNP-*hard*.

## 3.4 Discussion

The previous characterizations and complexity results lead to several questions: when is it worthwhile to preprocess a knowledge base by computing independence relations? How should these independence relations be computed? What is the level of generality of the definitions and results we gave in this section?

Many equivalent characterizations of formula-variable independence have been given in the literature, each of which could serve as a definition (Definition 4, Corollary 1, each of the statements (2), (3) and (4) in Proposition 7 and of the statements (2), (3) in Proposition 9), so one may wonder which one has to be used in practice.

In many papers referring explicitly to formula-variable independence, the prime implicant/cate characterization (Proposition 9) is used as a definition (Boutilier, 1994; Doherty et al., 1998). Generally speaking, this is not the cheapest way to compute the set of variables a formula depends on, since the size of $PI(\Sigma)$ is exponential in the size of $\Sigma$ in the worst case. This characterization is to be used in practice only if the syntactical form of $\Sigma$ is such that its prime implicants or prime implicates can be computed easily from it (this is the case for instance whenever $\Sigma$ is a Krom formula). Clearly, the cheapest way to compute formula-variable independence consists in using any of the equivalent formulations of Proposition 7, which all consist of validity tests.

Checking whether a formula $\Sigma$ is Var-independent from a variable $x$ is coNP-complete, which implies that simplifying a knowledge base by getting rid of redundant variables needs $|Var(\Sigma)|$ calls to an NP oracle, and is thus in $F\Delta_2^p$ and not below (unless $NP = coNP$). This may look paradoxical (and sometimes useless) to preliminarily compute several instances of a NP or coNP-hard independence problem to help solving a (single) instance of a NP or coNP-complete problem. However, this negative comment has a general scope only, and in many particular cases, this can prove quite efficient (indeed, even when $\Sigma$ has no particular syntactical form, the satisfiability or the unsatisfiability of $\Sigma_{x \leftarrow 0} \wedge \neg \Sigma_{x \leftarrow 1}$ may be particularly easy to check). Furthermore, if the knowledge base $\Sigma$ is to be queried many times, then the preprocessing phase consisting in Var-simplifying $\Sigma$ by ignoring useless variables is likely to be worthwhile.

## 4. Forgetting

In this section, we define what forgetting is, present some of its properties, and finally give some complexity results.

### 4.1 Definitions and Properties

A basic way to simplify a KB w.r.t. a set of literals or a set of variables consists in *forgetting* literals/variables in it. Beyond the simplification task, forgetting is a way to make a formula independent from literals/variables. Let us first start with literal forgetting:

**Definition 7 (literal forgetting)** *Let $\Sigma$ be a formula from $PROP_{PS}$ and $L$ be a subset of $L_{PS}$. $ForgetLit(\Sigma, L)$ is the formula inductively defined as follows:*

1. *$ForgetLit(\Sigma, \emptyset) = \Sigma$,*

2. *$ForgetLit(\Sigma, \{l\}) = \Sigma_{l \leftarrow 1} \vee (\neg l \wedge \Sigma)$,*

3. *$ForgetLit(\Sigma, \{l\} \cup L) = ForgetLit(ForgetLit(\Sigma, L), \{l\})$.*

This definition is sound since the ordering in which literals of $L$ are considered does not matter[4]. We can also prove that the definition above is equivalent to the one in which Point 2. is replaced by

$$ForgetLit(\Sigma, \{l\}) = \Sigma_{l \leftarrow 1} \vee (\neg l \wedge \Sigma_{l \leftarrow 0})$$

just because $\Sigma$ and $\Sigma_{l \leftarrow 0}$ are equivalent modulo $\neg l$.

Let us now give a semantical characterization of literal forgetting. If $L = \{l\}$, that is, $L$ is composed of a single literal, then forgetting $l$ from a formula $\Sigma$ amounts to introducing the model $Force(\omega, \neg l)$ for each model $\omega$ of $\Sigma$ such that $\omega \models l$.

**Proposition 14** *The set of models of $ForgetLit(\Sigma, \{l\})$ can be expressed as:*

$$
\begin{aligned}
Mod(ForgetLit(\Sigma, \{l\})) &= Mod(\Sigma) \cup \{Force(\omega, \neg l) \mid \omega \models \Sigma\} \\
&= \{\omega \mid Force(\omega, l) \models \Sigma\}
\end{aligned}
$$

A similar statement can be given for the case in which $L$ is composed of more than one literal. In this case, for each model $\omega$ of $\Sigma$, we can force any subset of literals $L_1 \subseteq L$ such that $\omega \models L_1$ to assume the value false.

**Proposition 15** *The set of models of $ForgetLit(\Sigma, L)$ can be expressed as:*

$$Mod(ForgetLit(\Sigma, L)) = \{\omega \mid Force(\omega, L_1) \models \Sigma \text{ where } L_1 \subseteq L\}$$

As a corollary, we obtain the following properties of literal forgetting:

**Corollary 2** *Let $\Sigma$, $\Phi$ be formulas from $PROP_{PS}$ and $L_1, L_2 \subseteq L_{PS}$.*

- $\Sigma \models ForgetLit(\Sigma, L_1)$ *holds.*

- *If $\Sigma \models \Phi$ holds, then $ForgetLit(\Sigma, L_1) \models ForgetLit(\Phi, L_1)$ holds as well.*

- *If $L_1 \subseteq L_2$ holds, then $ForgetLit(\Sigma, L_1) \models ForgetLit(\Sigma, L_2)$ holds as well.*

Let now consider an example.

**Example 11** *Let $\Sigma = (\neg a \vee b) \wedge (a \vee c)$. We have $ForgetLit(\Sigma, \{\neg a\}) \equiv (a \vee c) \wedge (b \vee c)$.*

The key proposition for the notion of forgetting is the following one:

---

4. The proof of this statement is almost straightforward: let $l_1$, $l_2$ be two literals;

1. if $l_1 \neq l_2$ and $l_1 \neq \neg l_2$ then $ForgetLit(ForgetLit(\Sigma, l_1), l_2) = \Sigma_{l_1 \leftarrow 1, l_2 \leftarrow 1} \vee (\neg l_1 \wedge \Sigma)_{l_2 \leftarrow 1} \vee (\neg l_2 \wedge \Sigma_{l_1 \leftarrow 1}) \vee (\neg l_1 \wedge \neg l_2 \wedge \Sigma) \equiv \Sigma_{l_1 \leftarrow 1, l_2 \leftarrow 1} \vee (\neg l_1 \wedge \Sigma_{l_2 \leftarrow 1}) \vee (\neg l_2 \wedge \Sigma_{l_1 \leftarrow 1}) \vee (\neg l_1 \wedge \neg l_2 \wedge \Sigma)$ is symmetric in $l_1$ and $l_2$;

2. if $l_2 = \neg l_1 = \neg l$ then $ForgetLit(ForgetLit(\Sigma, l), \neg l) = (\Sigma_{l \leftarrow 1} \vee (\neg l \wedge \Sigma_{l \leftarrow 0}))_{l \leftarrow 0} \vee (l \wedge (\Sigma_{l \leftarrow 1} \vee (\neg l \wedge \Sigma_{l \leftarrow 0}))_{l \leftarrow 0}) \equiv \Sigma_{l \leftarrow 1} \vee \Sigma_{l \leftarrow 0}$ is symmetric in $l$ and $\neg l$;

3. the case $l_1 = l_2$ is trivial.

**Proposition 16** *Let $\Sigma$ be a formula from $PROP_{PS}$ and $L \subseteq L_{PS}$. $ForgetLit(\Sigma, L)$ is the logically strongest consequence of $\Sigma$ that is Lit-independent from $L$ (up to logical equivalence).*

The following immediate consequence of Proposition 16 establishes strong relationships between literal forgetting and Lit-independence:

**Corollary 3** *Let $\Sigma$ be a formula from $PROP_{PS}$ and $L \subseteq L_{PS}$. $\Sigma$ is Lit-independent from $L$ if and only if $\Sigma \equiv ForgetLit(\Sigma, L)$ holds.*

The following one gives an immediate application of literal forgetting:

**Corollary 4** *If a formula $\varphi$ is Lit-independent from $L$, then $\Sigma \models \varphi$ holds if and only if $ForgetLit(\Sigma, L) \models \varphi$.*

This result proves that forgetting literals from $L$ does not affect entailment of formulas that are Lit-independent from $L$. This is in some sense analogous to the concept of filtration in modal logics (Goldblatt, 1987); indeed, if we are interested in knowing whether $\Sigma \models \varphi$ only for formulas $\varphi$ that are Lit-independent from $L$, then the literals of $L$ can be forgotten in $\Sigma$.

Let us now investigate the computation of $ForgetLit(\Sigma, L)$. Let us first consider DNF formulas $\Sigma$. Forgetting literals within DNF formulas is a computationally easy task. On the one hand, forgetting literals within a disjunctive formula comes down to forgetting them in every disjunct:

**Proposition 17** *Let $\Sigma$, $\Phi$ be two formulas from $PROP_{PS}$ and $L \subseteq L_{PS}$.*

$$ForgetLit(\Sigma \vee \Phi, L) \equiv ForgetLit(\Sigma, L) \vee ForgetLit(\Phi, L).$$

On the other hand, forgetting literals within a consistent term simply consists in removing them from the term:

**Proposition 18** *Let $\gamma$ be a consistent term from $PROP_{PS}$ (viewed as the set of its literals) and $L \subseteq L_{PS}$. $ForgetLit(\gamma, L) \equiv \bigwedge_{l \in \gamma \setminus L} l$.*

Combining the two previous propositions shows how literals can be forgotten from a DNF formula $\Sigma$ in polynomial time. It is sufficient to delete every literal of $L$ from each disjunct of $\Sigma$ (if one of the disjuncts becomes empty then $ForgetLit(\Sigma, L) \equiv true$).

Things are more complicated for conjunctive formulas $\Sigma \wedge \Phi$. Especially, there is no result similar to Proposition 17 for conjunctive formulas. While $ForgetLit(\Sigma, L) \wedge ForgetLit(\Phi, L)$ is a logical consequence of $ForgetLit(\Sigma \wedge \Phi, L)$ (see Corollary 2), the converse does not hold in the general case.

**Example 12** *Let $\Sigma = a$, $\Phi = \neg a$, and $L = \{a\}$. Since $a \not\models \Phi$, we have $ForgetLit(\Phi, L) \equiv \Phi$. Since $ForgetLit(\Sigma, L)$ is valid, we have $(ForgetLit(\Sigma, L) \wedge ForgetLit(\Phi, L)) \equiv \neg a$. Since $\Sigma \wedge \Phi$ is inconsistent, $ForgetLit(\Sigma \wedge \Phi, L)$ is inconsistent as well.*

Clearly enough, any non-valid clause $\delta$ is Lit-independent from $L$ if and only if $Lit(\delta) \cap L = \emptyset$. Since the conjunction of two formulas that are Lit-independent from $L$ is Lit-independent from $L$ (see Proposition 2), and since every formula is equivalent to a CNF formula, Proposition 16 shows $ForgetLit(\Sigma, L)$ equivalent to the set of all clauses $\delta$ that are entailed by $\Sigma$ and are from $\{\delta$ clause $\mid Lit(\delta) \cap L = \emptyset\}$. Because $\{\delta$ clause $\mid Lit(\delta) \cap L = \emptyset\}$ is closed under subsumption (i.e., it is a stable production field), it is possible to take advantage of consequence finding algorithms (see Marquis, 2000), to derive a CNF representation of $ForgetLit(\Sigma, L)$. Especially, in the case where $\Sigma$ is a CNF formula, resolution-based consequence finding algorithms like those reported in (Inoue, 1992) or (del Val, 1999) can be used; this is not very surprising since resolution is nothing but *variable elimination*.

In contrast to the disjunctive formulas situation, there is no guarantee that such consequence-finding algorithms run in time polynomial in the input size when the input is a conjunctive formula (otherwise, as explained in the following, we would have $\mathsf{P} = \mathsf{NP}$). Nevertheless, forgetting literals within a conjunctive formula $\Sigma$ can be easy in some restricted cases, especially when $\Sigma$ is given by the set of its prime implicates; in this situation, it is sufficient to give up those clauses containing a literal from $L$.

**Proposition 19** *Let $\Sigma$ be a formula from $PROP_{PS}$ and $L \subseteq L_{PS}$.*

$$IP(ForgetLit(\Sigma, L)) = \{\delta \mid \delta \in IP(\Sigma) \text{ and } Lit(\delta) \cap L = \emptyset\}.$$

The notion of literal forgetting generalizes the notion of variable elimination from propositional logic (that was already known by Boole as elimination of middle terms and has been generalized to the first-order case in a more recent past by Lin & Reiter, 1994). Indeed, variable elimination is about *variable forgetting*, i.e., the one achieved not considering the literals signs:

**Definition 8 (variable forgetting)** *Let $\Sigma$ be a formula from $PROP_{PS}$ and let $V$ be a subset of $PS$. $ForgetVar(\Sigma, V)$ is the formula inductively defined as follows:*

- $ForgetVar(\Sigma, \emptyset) = \Sigma$,

- $ForgetVar(\Sigma, \{x\}) = \Sigma_{x \leftarrow 1} \vee \Sigma_{x \leftarrow 0}$,

- $ForgetVar(\Sigma, \{x\} \cup V) = ForgetVar(ForgetVar(\Sigma, V), \{x\})$.

**Example 13** *Let $\Sigma = (\neg a \vee b) \wedge (a \vee c)$. We have $ForgetVar(\Sigma, \{a\}) \equiv (b \vee c)$.*

As a direct consequence of the definition, $ForgetVar(\Sigma, \{x_1, ..., x_n\})$ is equivalent to the quantified boolean formula (usually with free variables!) denoted $\exists x_1 \ldots \exists x_n \Sigma$.

Clearly enough, forgetting a variable $x$ amounts to forgetting both the literals $x$ and $\neg x$.

**Proposition 20** *Let $\Sigma$ be a formula from $PROP_{PS}$ and $V \subseteq PS$. We have*

$$ForgetVar(\Sigma, V) \equiv ForgetLit(\Sigma, L_V)$$

This result, together with the previous results on literal forgetting, gives us the following corollaries for variable forgetting:

**Corollary 5**

$$Mod(ForgetVar(\Sigma, \{x\})) \quad = \quad Mod(\Sigma) \cup \{Switch(\omega, x) \mid \omega \models \Sigma\}.$$

**Corollary 6** *Let $\Sigma$ be a formula and $V \subseteq PS$. $ForgetVar(\Sigma, V)$ is the logically strongest consequence of $\Sigma$ that is Var-independent from $V$ (up to logical equivalence).*

**Corollary 7** *If a formula $\varphi$ is Var-independent from $V$, then $\Sigma \models \varphi$ if and only if $ForgetVar(\Sigma, V) \models \varphi$.*

A consequence of the latter result is that forgetting variables is useful when only a subset of variables are really used in the queries. Thus, if $\Sigma$ represents some pieces of knowledge about a scenario of interest, and we are interested in knowing whether a fact $\varphi$ is true in the scenario, the logical operation to do is to query whether $\Sigma \models \varphi$. Now, if the possible facts $\varphi$ we are interested in do not involve some variables $V$, then these variables can be forgotten from $\Sigma$, as querying whether $\varphi$ is implied can be done on $ForgetVar(\Sigma, V)$ instead of $\Sigma$.

Through the previous proposition, some algorithms for forgetting variables can be easily derived from algorithms for forgetting literals (some of them have been sketched before). Specifically, polynomial time algorithms for forgetting variables within a DNF formula or a formula given by the set of its prime implicates can be obtained. Other tractable classes of propositional formulas for variable forgetting exist. For instance, taking advantage of the fact that $ForgetVar(\Sigma \wedge \Phi, V) \equiv ForgetVar(\Sigma, V) \wedge ForgetVar(\Phi, V)$ whenever $Var(\Sigma) \cap Var(\Phi) = \emptyset$ holds, Darwiche (1999) showed that variable forgetting in a formula $\Sigma$ can be done in linear time as soon as $\Sigma$ is in *Decomposable Negation Normal Form* (DNNF), i.e., a (DAG)NNF formula in which the conjuncts of any conjunctive subformula do not share any variable. Interestingly, the DNNF fragment of propositional logic is strictly more succinct than the DNF one (especially, some DNNF formulas only admit exponentially-large equivalent DNF formulas) (Darwiche & Marquis, 1999).

In the general case, just as for the literal situation, there is no way to forget efficiently (i.e., in polynomial time) a set of variables within a formula (unless $P = NP$). Nevertheless, the following decomposition property can be helpful in some situations (actually, it is heavily exploited in Kohlas et al., 1999).

**Proposition 21** *Let $\Sigma$, $\Phi$ be two formulas from $PROP_{PS}$, and $V$ be a subset of $PS$. If $V \not\hookrightarrow_-^+ \Sigma$, then $ForgetVar(\Sigma \wedge \Phi, V) \equiv \Sigma \wedge ForgetVar(\Phi, V)$.*

Note that the corresponding property for literal forgetting does not hold (as a previous example shows).

Forgetting literals or variables proves helpful in various settings (we already sketched some of them in the introduction). For instance, minimal model inference (or circumscription McCarthy, 1986) can be expressed using literal forgetting (but not directly using variable forgetting, which shows the interest of the more general form we introduced). Indeed, it is well-known that closed world inference from a knowledge base $\Sigma$ can be logically characterized as classical entailment from $\Sigma$ completed with some assumptions. In the circumscription framework (McCarthy, 1986), given a partition $\langle P, Q, Z \rangle$ of $PS$, such assumptions are the negations of the formulas $\alpha$ s.t. $\alpha$ does not contain any variable from $Z$ and for every clause $\gamma$ containing only positive literals built up from $P$ and literals built up from $Q$, if $\Sigma \not\models \gamma$ holds, then $\Sigma \not\models \gamma \vee \alpha$ holds as well. Equivalently, $\neg\alpha$ is considered a reasonable assumption whenever it is (syntactically) Var-independent from $Z$ and expanding $\Sigma$ with it does not modify what is already known about $L_P^+ \cup L_Q$. Clearly enough, the signs of literals from $L_P$ really matter here. Using our previous notations, $\neg\alpha$ is assumed if and only if $Z \not\mapsto_-^+ \neg\alpha$ and $\Sigma \equiv_{L_P^+ \cup L_Q} \Sigma \wedge \neg\alpha$. As a consequence, we derive the following characterization of circumscription:

**Proposition 22** *Let $\Sigma$, $\Phi$ be two formulas from $PROP_{PS}$, and $P$, $Q$, $Z$ be three disjoint sets of variables from $PS$ (such that $Var(\Sigma) \cup Var(\Phi) \subseteq P \cup Q \cup Z$). It holds:*

- *If $\Phi$ does not contain any variable from $Z$*

$$CIRC(\Sigma, \langle P, Q, Z \rangle) \models \Phi$$
$$\textit{if and only if}$$
$$\Sigma \models ForgetLit(\Sigma \wedge \Phi, L_P^- \cup L_Z)$$

- *In the general case:*

$$CIRC(\Sigma, \langle P, Q, Z \rangle) \models \Phi$$
$$\textit{if and only if}$$
$$\Sigma \models ForgetLit(\Sigma \wedge \neg ForgetLit(\Sigma \wedge \neg\Phi, L_Z \cup L_P^-), L_Z \cup L_P^-)$$

*where $CIRC$ is circumscription as defined in (McCarthy, 1986).*

Similar characterizations can be derived for the other forms of closed world reasoning pointed out so far.

Forgetting also is a central concept when we are concerned with query answering w.r.t. a restricted target language. Indeed, in many problems, there is a set of variables for which we are not interested in their truth value (so we can forget them). For instance, in the SATPLAN framework by Kautz, McAllester, and Selman (1996), compiling away fluents or actions amounts to forgetting variables. Since the only variables we are really interested in within a given set of clauses representing a planning problem instance are those representing the plans, we can compile away any other variable, if this does not introduce an increase of size of the resulting formula. Another situation where such a forgetting naturally occurs is model-based diagnosis (Reiter, 1987); compiling away every variable except the abnormality ones does not remove any piece of information required to compute the conflicts and the diagnoses of a system. Thus, Darwiche (1998) shows how both the set of conflicts and the set of consistency-based diagnoses of a system is characterized by the formula obtained by forgetting every variable except the abnormality ones in the conjunction of the system description and the available observations. Provided that the system description has first been turned into DNNF, forgetting can be achieved in linear time and diagnoses containing a minimal number of faulty components can be enumerated in (output) polynomial time. Interestingly, this work shows that the diagnosis task does not require the (usually expensive) computation of prime implicates/implicants to be achieved (actually, computing prime implicates/implicants is just a *way* to achieve variable forgetting and not a *goal* in consistency-based diagnosis). Forgetting every variable from a formula allows for consistency checking since $\Sigma$ is consistent if and only if $ForgetVar(\Sigma, Var(\Sigma))$ is consistent. The well-known Davis and Putnam algorithm for satisfiability testing (Davis & Putnam, 1960) (recently revisited by Dechter and Rish (1994)under the name directional resolution) basically consists in computing a clausal representation of $ForgetVar(\Sigma, Var(\Sigma))$ from a CNF $\Sigma$ using resolution; if the empty clause is not generated, then $\Sigma$ is consistent and the converse also holds.

Forgetting can also be used as a key concept in order to organize knowledge so as to replace one global inference into a number of local inferences as shown (among others) by Kohlas et al. (1999) and Amir and McIlraith (2000), McIlraith and Amir (2001). Loosely speaking, such approaches rely on the idea that exploiting all the pieces of information given in a knowledge base is typically not required for query answering. Focusing on what is relevant to the query is sufficient. While such techniques do not lower the complexity of inference from the theoretical side, they can lead to significant practical improvements. For instance, assume that $\Sigma$ consists of three formulas $\Phi_1$, $\Phi_2$, and $\Phi_3$. For any query $\Psi$, let $V_\Psi = (\bigcup_{i=1}^{3} Var(\Phi_i)) \setminus Var(\Psi)$. We have $\Sigma \models \Psi$ if and only if $ForgetVar(\bigwedge_{i=1}^{3} \Phi_i, V_\Psi) \models \Psi$. If $Var(\Phi_3) \cap (\bigcup_{i=1}^{2} Var(\Phi_i)) = \emptyset$, this amounts to test *independently* whether $ForgetVar(\bigwedge_{i=1}^{2} \Phi_i, V_\Psi) \models \Psi$ holds or $ForgetVar(\Phi_3, V_\Psi) \models \Psi$ holds. This way, one global inference is replaced by two local inferences. Now, $ForgetVar(\Phi_1 \wedge \Phi_2, V_\Psi)$ is equivalent to $ForgetVar(\Phi_1 \wedge ForgetVar(\Phi_2, V_\Psi \cap (Var(\Phi_2) \setminus Var(\Phi_1))), V_\Psi)$. Accordingly, every variable from $\Phi_2$ that is not a variable of $\Phi_1$ or a variable of $\Psi$ can be forgotten first within $\Phi_2$ because it gives no information relevant to the query; thus, only a few pieces of knowledge have to be "propagated" from $\Phi_2$ to $\Phi_1$ before answering the query, and forgetting allows for characterizing them exactly.

As evoked in the introduction, another scenario in which forgetting is useful is that of belief update. Indeed, there are many formalizations of belief update that are based on a form of variable forgetting. The basic scenario is the following one: we have a formula $\Sigma$ that represents our knowledge; there are some changes in the world, and what we know is that after them a formula $\varphi$ becomes true. The simplest way to deal with the update is to assume that $\varphi$ represents all we know about the truth value of the variables in $\varphi$. As a result, we have to "forget" from $\Sigma$ the value of the variables in $Var(\varphi)$. There are different formalizations of this schema, based on whether formula $\varphi$ is considered to carry information about variables it mentions (Winslett, 1990) or only on the variables it depends on (Hegner, 1987), or also on variables related to dependent variables via a dependence function (Herzig, 1996). This kind of update schema, while less known than the Possible Models Approach by Winslett (1990), has proved to be suited for reasoning about actions (Doherty et al., 1998; Herzig & Rifi, 1999). Furthermore, the possibility to forget literals (and not variables) is also valuable in this framework to take account for persistent information, as shown recently by some of us (Herzig et al., 2001), since the polarity of information is often significant. For instance, while forgetting the fluent *alive* from a knowledge base is not problematic, forgetting the persistent fluent $\neg alive$ would surely be inadequate.

Forgetting can also be used to characterize a dependence relation called definability (Lang & Marquis, 1998b) as well as the set of strongest necessary (resp. weakest sufficient) conditions of a propositional variable on a set $V$ of variables given a theory $\Sigma$ (Lin, 2000; Doherty, Lukaszewicz, & Szalas, 2001). As shown in (Lang & Marquis, 1998b; Lin, 2000; Doherty et al., 2001), all these notions have many applications in various AI fields, including hypothesis discrimination, agent communication, theory approximation and abduction.

Finally, based on literal and variable forgetting, valuable equivalence relations over formulas can also be defined:

**Definition 9 (Lit-equivalence, Var-equivalence)** *Let $\Sigma$, $\Phi$ be two formulas from $PROP_{PS}$, $L$ be a subset of $L_{PS}$, and $V$ be a subset of $PS$.*

- *$\Sigma$ and $\Phi$ are said to be Lit-equivalent given $L$, denoted $\Sigma \equiv_L \Phi$, if and only if $ForgetLit(\Sigma, Lit(\Sigma) \setminus L) \equiv ForgetLit(\Phi, Lit(\Phi) \setminus L)$.*

- *$\Sigma$ and $\Phi$ are said to be Var-equivalent given $V$, denoted $\Sigma \equiv_V \Phi$, if and only if $ForgetVar(\Sigma, Var(\Sigma) \setminus V) \equiv ForgetVar(\Phi, Var(\Phi) \setminus V)$.*

**Example 14** *Let $\Sigma = (a \Rightarrow b) \wedge (b \Rightarrow c)$ and $\Phi = (a \Rightarrow d) \wedge (d \Rightarrow c)$. Let $L = \{\neg a, c\}$. $\Sigma$ and $\Phi$ are Lit-equivalent given $L$.*

Such equivalence relations capture some forms of dependence between formulas. They are useful in the various situations where it is required to formally characterize the fact that two knowledge bases share some theorems. Thus, two formulas are Lit-equivalent given $L$ whenever every clause containing literals from $L$ only is a logical consequence of the first formula if and only if it is a logical consequence of the second formula. In the same vein, two Var-equivalent formulas given $V$ have the same clausal consequences built up from $V$. Clearly, Lit-equivalence is more fine-grained than Var-equivalence in the sense that two formulas Var-equivalent given $V$ are also Lit-equivalent given $L = L_V$, the set of literals built upon $V$, but the converse does not hold in the general case. Some applications are related to knowledge approximation ($\Phi$ is a correct approximation of $\Sigma$ over $L$ if and only if $\Phi$ and $\Sigma$ are Lit-equivalent given $L$), normalization (turning a formula $\Sigma$ into a CNF $\Phi$ by introducing new symbols is acceptable as long as the two formulas are equivalent over the original language, i.e., $\Sigma$ and $\Phi$ are Var-equivalent given $V = Var(\Sigma)$), and so on.

## 4.2 Complexity Results

It is quite easy to prove that forgetting is a computationally expensive operation in the general case. Indeed, since a formula $\Sigma$ is consistent if and only if $ForgetLit(\Sigma, Lit(\Sigma))$ is consistent and since the latter formula is Lit-independent from every literal (i.e., it is equivalent to *true* or equivalent to *false*), there is no way to compute a formula $\Phi$ equivalent to $ForgetLit(\Sigma, L)$ in polynomial time, unless $\mathsf{P} = \mathsf{NP}$. Actually, we can derive the more constraining result, showing that the *size* of any formula equivalent to $ForgetLit(\Sigma, L)$ may be superpolynomially larger than the size of $\Sigma$.

**Proposition 23** *Let $\Sigma$ be a formula from $PROP_{PS}$ and let $L$ be a finite subset of $L_{PS}$. In the general case, there is no propositional formula $\Phi$ equivalent to $ForgetLit(\Sigma, L)$ s.t.*

*the size of $\Phi$ is polynomially bounded in $|\Sigma| + |L|$, unless $\mathsf{NP} \cap \mathsf{coNP} \subseteq \mathsf{P/poly}$ (which is considered unlikely in complexity theory).*

This result shows that computing an explicit representation of $ForgetLit(\Sigma, L)$ under the form of a propositional formula is hard, even in a compilation-based approach where the time needed to derive such a formula is neglected.

Finally, we have also derived:

**Proposition 24 (complexity of Lit/Var-equivalence)**
LIT-EQUIVALENCE *and* VAR-EQUIVALENCE *are* $\Pi_2^p$-*complete.*

## 5. Influenceability, Relevance, and Strict Relevance

In this section, we show that several notions of dependence introduced in the literature are equivalent to, or can be expressed in terms of, semantical independence. In particular, we show that Boutilier's definition of influenceability (Boutilier, 1994) is in fact equivalent to (semantical) FV-dependence. The definition of relevance as given by Lakemeyer (1997) can also be proved to be equivalent to FV-dependence. One of the two definitions given by Lakemeyer (1997) for strict relevance can also be expressed in terms of FV-dependence. These results allow for finding the complexity of all these forms of dependence as a direct corollary to the complexity results reported in the previous section. For the sake of completeness, we also give the complexity of the original definition of strict relevance (Lakemeyer, 1995) (which is not directly related to FV-dependence), which turns out to be computationally simpler than the subsequent definition given by Lakemeyer (1997).

### 5.1 Influenceability

Boutilier (1994) introduces a notion of influenceability. Roughly speaking, a formula $\Sigma$ is influenceable from a set of variables $V$ if there exists a scenario in which the truth value of $\Sigma$ depends on the value of the variables in $V$. This idea can be formalized as follows.

**Definition 10 (influenceability)** *Let $\Sigma$ be a formula from $PROP_{PS}$ and $V$ a subset of $PS$. $\Sigma$ is* influenceable *from $V$ if and only if there exists a $PS \setminus V$-world $\omega$, and two $V$-worlds $\omega_1$ and $\omega_2$ s.t. $\omega \wedge \omega_1 \models \Sigma$ and $\omega \wedge \omega_2 \models \neg\Sigma$ hold.*

In other words, there is a scenario $\omega$ in which the formula $\Sigma$ can be true or false, depending on the value of the variables in $V$. While influenceability looks different from the definitions given in this paper, it can be shown that in fact influenceability coincides with FV-dependence.

**Proposition 25** *Let $\Sigma$ be a formula from $PROP_{PS}$ and $V$ a subset of $PS$. $\Sigma$ is influenceable from $V$ if and only if $\Sigma$ is Var-dependent on $V$.*

As a consequence, a model-theoretic characterization of influenceability can be easily derived from the one for FV-independence. The complexity of influenceability is an easy corollary to this property: INFLUENCEABILITY is NP-complete.

## 5.2 Relevance

Lakemeyer (1995, 1997) introduces several forms of relevance. We show how these forms of relevance are strongly related to FV-independence. We also complete the results given in (Lakemeyer, 1997), by exhibiting the computational complexity of each form of relevance introduced in (Lakemeyer, 1997).

### 5.2.1 RELEVANCE OF A FORMULA TO A SUBJECT MATTER

Lakemeyer's notion of relevance of a formula to a subject matter can be defined in terms of prime implicates of the formula, as follows (see Definition 9 in Lakemeyer, 1997):

**Definition 11 (relevance to a subject matter)** *Let $\Sigma$ be a formula from $PROP_{PS}$ and $V$ a subset of $PS$. $\Sigma$ is relevant to $V$ if and only if there exists a prime implicate of $\Sigma$ mentioning a variable from $V$.*

**Example 15** *Let $\Sigma = (a \wedge c)$ and $V = \{a, b\}$. $\Sigma$ is relevant to $V$.*

As a consequence, Lakemeyer's notion of irrelevance of a formula to a subject matter coincides with FV-independence.

**Proposition 26** *Let $\Sigma$ be a formula from $PROP_{PS}$ and $V$ a subset of $PS$. $\Sigma$ is relevant to $V$ if and only if $\Sigma$ is Var-dependent on $V$.*

Thus, the model-theoretic characterization of FV-independence also applies to irrelevance of a formula to a subject matter. We also have that the irrelevance of a formula to a subject matter coincides with Boutilier's definition of influenceability. Finally, the above proposition allows for an easy proof of complexity for relevance, namely, RELEVANCE OF A FORMULA TO A SUBJECT MATTER is NP-complete.

### 5.2.2 STRICT RELEVANCE OF A FORMULA TO A SUBJECT MATTER

Lakemeyer has introduced two forms of *strict relevance*. The (chronologically) first one has been given in (Lakemeyer, 1995), as follows.

**Definition 12 (strict relevance to a subject matter, Lakemeyer, 1995)** *Let $\Sigma$ be a formula from $PROP_{PS}$ and $V$ a subset of $PS$. $\Sigma$ is strictly relevant to $V$ if and only if every prime implicate of $\Sigma$ contains a variable from $V$.*

Lakemeyer has also introduced another notion of strict relevance (Lakemeyer, 1997), more demanding than the original one. Here we consider an equivalent definition.

**Definition 13 (strict relevance to a subject matter, Lakemeyer, 1997)** *Let $\Sigma$ be a formula from $PROP_{PS}$ and $V$ a subset of $PS$. $\Sigma$ is strictly relevant to $V$ if and only if there exists a prime implicate of $\Sigma$ mentioning a variable from $V$, and every prime implicate of $\Sigma$ mentions only variables from $V$.*

Both definitions prevent tautologies and contradictory formulas from being strictly relevant to any set of variables. The basic difference between these two definitions is that in the first one we want that every prime implicate of $\Sigma$ contains *at least* a variable from $V$, while in the second case we impose that every prime implicate of $\Sigma$ must contain *only* variables from $V$[5]. As the following example shows, there are formulas for which the two definitions of strict relevance do not coincide.

**Example 16** *Let* $\Sigma = (a \lor b)$ *and* $V = \{a\}$. *There is only one prime implicate of* $\Sigma$, *namely* $a \lor b$. *Since it contains at least a variable of* $V$, *it follows that* $\Sigma$ *is strictly relevant to* $V$ *w.r.t. (Lakemeyer, 1995). However, since the prime implicate* $a \lor b$ *is not composed only of variables of* $V$ *(because* $b \notin V$*), it follows that* $\Sigma$ *is not strictly relevant to* $V$ *w.r.t. (Lakemeyer, 1997).*

Through FV-independence, we can derive an alternative characterization of the notion of *strict relevance* introduced by Lakemeyer (1997). Indeed, as a straightforward consequence of the definition, we have:

**Proposition 27** *Let* $\Sigma$ *be a formula from* $PROP_{PS}$ *and* $V$ *a subset of* $PS$. $\Sigma$ *is strictly relevant to* $V$ *if and only if* $\Sigma$ *is Var-dependent on* $V$ *and Var-independent from* $Var(\Sigma) \setminus V$.

We have identified the complexity of both definitions of strict relevance, and they turn out to be different: the first definition is harder than the second one.

**Proposition 28 (complexity of strict relevance)**
*(1)* STRICT RELEVANCE OF A FORMULA TO A SUBJECT MATTER *(Lakemeyer, 1995) is* $\Pi_2^p$*-complete.*
*(2)* STRICT RELEVANCE OF A FORMULA TO A SUBJECT MATTER *(Lakemeyer, 1997) is* $\mathsf{BH}_2$*-complete.*

These complexity results improve Theorem 50 from (Lakemeyer, 1997), which only points out the NP-hardness of irrelevance and strict irrelevance as defined in (Lakemeyer, 1997).

## 6. Other Related Work and Further Extensions

In this section, we first discuss other related work, then some possible extensions of the notions and results we have presented before.

### 6.1 Other Related Work

As already evoked, independence has been considered under various forms in various AI fields.

---

5. Strict relevance as in (Lakemeyer, 1997) could also be shown to be strongly related to controllability (Boutilier, 1994; Lang & Marquis, 1998b).

### 6.1.1 Independence in propositional logic

There are other forms of independence in propositional logic that we have not considered in this article, especially, definability, controllability (Lang & Marquis, 1998b) as well as conditional independence (Darwiche, 1997). If $X$, $Y$ and $Z$ are three disjoint sets of variables and $\Sigma$ is a knowledge base then $X$ and $Y$ are conditionally independent w.r.t. $Z$ knowing $\Sigma$ if and only if for any $Z$-world $\omega_Z$, once we know $\omega_Z$ (and $\Sigma$), learning something about $X$ cannot make us learn anything new about $Y$ (and *vice versa*). The computational issues pertaining to conditional independence and to stronger notions as well as related notions such as relevance between subject matters (Lakemeyer, 1997) and novelty (Greiner & Genesereth, 1983), have been extensively studied in a companion paper (Lang, Liberatore, & Marquis, 2002).

### 6.1.2 Implicit and explicit dependence

Several approaches to belief change make use of a explicit dependence relation, which means that it is part of the input (while ours is implicit, i.e., derived from the input). Thus, computing independence relations from a knowledge base can be seen as an upstream task enabling us to specify the "core" (minimal) independence relation upon which the belief change operator is based; this core independence relation can then be completed by specifying explicitly some additional dependencies using knowledge about the domain. Such an approach has been proposed for belief revision in (Fariñas del Cerro & Herzig, 1996), for belief update in (Marquis, 1994) and (Herzig, 1996) and for reasoning about action in (Herzig & Rifi, 1999).

### 6.1.3 Independence and contexts

Contextual reasoning (Ghidini & Giunchiglia, 2001) has been introduced for formalizing domains in which knowledge can naturally be divided into parts (contexts). Each context is characterized by its own language and alphabet. The knowledge base of a context contains what is relevant to a part of the domain. However, it is not guaranteed that the different parts of the domain do not interact, so inference in one context may be affected by the knowledge of some other context.

The main difference between contextual reasoning and independence is that the latter is a study of the relevance relation that can be drawn from a "flat" (i.e., not divided into contexts) knowledge base; whereas contextual reasoning is on knowledge about specific contexts, that is, knowledge is expressed by specifying which context it refers to. In other words, the relevance relation is a result of reasoning about knowledge in studying dependency; on the other hand, it is one of the data that has to be provided for reasoning about contexts.

### 6.1.4 The definition of irrelevance by Levy et al.

The definition of irrelevance given by Levy, Fikes, and Sagiv (1997) aims at establishing which facts of a knowledge base are irrelevant to the derivation of a query. In particular, they consider a first-order logic with no function symbols and a set of inference rules. A knowledge base is a set of closed formulas (formulas with no free variables). Derivation of a query (another closed formula) is obtained by applying the inference rules to the knowledge base and the logical axioms of the theory.

A formula $\phi$ of the knowledge base is irrelevant to the derivation of another formula $\psi$ if $\phi$ does not "participate" to the process of inferring $\psi$ from the knowledge base. For example, in the knowledge base $\{A(1), C(2), A(X) \Rightarrow B(X)\}$, it is clear that $A(1)$ is relevant to $B(1)$, while $C(2)$ is not.

This definition becomes complicated when more complex scenarios are considered. Namely, Levy et al. consider three different "coordinates": first, whether all derivations are considered or just one; second, whether we consider all derivations or just minimal ones; third, whether we consider membership to the proof or just derivability from the formulas that compose the proof (in this case, we have four possible choices).

Besides the fact that this notion of irrelevance is based on first-order logic, there are other, more substantial, differences between it and the ideas investigated in this paper. First, it is a relation between two formulas, given a background knowledge base. As such, it is more related to other notions of relevance in the literature (Lang & Marquis, 1998a). Second, it is based on the concept of proof, which is something not completely dependent on the semantics. For example, replacing the usual modus ponens with the strange inference rule $\alpha, \beta, \alpha \Rightarrow \gamma \to \gamma$, which does not change the semantics of the logics, then the formula $C(2)$ of the knowledge base above becomes magically relevant to $B(1)$. This is perfectly reasonable in the approach by Levy et al., where improving efficiency using a specific proof theory is the aim.

## 6.2 Extending our Notions and Results

The notions and results presented in this paper can be extended in several directions.

### 6.2.1 NUMERICAL AND ORDINAL NOTIONS OF FORMULA-VARIABLE INDEPENDENCE; FORGETTING IN MORE GENERAL STRUCTURES

It may be worth wondering about how the notions of FL-independence and FV-independence can be generalized to the case where the knowledge base is not a mere propositional formula but a probability distribution over $\Omega_{PS}$, or an incomplete probability (or equivalently a set of probability distributions), or a stratified knowledge base, or any other ordinal or numerical structure.

First, let us notice that some of our characterizations lead to quite intuitive and general ideas. To take the case of FV-independence, we have shown that the following three statements are equivalent when $\Sigma$ is a propositional knowledge base:

(a) $\Sigma$ does not tell anything about $x$, in any context;

(b) $\Sigma$ can be rewritten equivalently in a formula $\Sigma'$ in which $x$ does not appear (Definition 4);

(c) for any two interpretations $\omega$ and $\omega'$ that differ only in the value given to $x$, the status of $\omega$ with respect to $\Sigma$ (i.e., model or countermodel) is the same as that of $\omega'$ (Corollary 1).

As to variable forgetting:

(d) $ForgetVar(\Sigma, V)$ is the most general consequence of $\Sigma$ that is Var-independent from $V$ (Corollary 16).

Now, these definitions have a sufficient level of generality to be extended to the case where the knowledge base $\Sigma$ is replaced by another structure.

Thus, some of us have extended the notion of Var-independence (characterized using (b)) and variable forgetting to ordinal conditional functions (represented as stratified bases) (Lang, Marquis, & Williams, 2001). The basic purpose was to extend the "forget and expand" approach at work for updating "flat" knowledge bases (as briefly discussed in Section 4) to updating more sophisticated epistemic states.

The case of incomplete probabilities is also interesting since generalizing (a), (b) and (c) will lead to the same intuitive notion [6]. As to variable forgetting, it is not hard to notice that it corresponds to the well-known notion of *marginalization*.

### 6.2.2 Formula-variable independence in nonclassical (propositional) logics

The definition of FL- and FV-independence ($\Sigma$ can be rewritten equivalently in a formula $\Sigma'$ in which $l$ (resp. $x$) does not appear) is quite general, in the sense that the logical language *and/or* the consequence relation (and thus the notion of logical equivalence) may vary, which enables us to draw from this principle notions of FL- and FV-independence (and from then on, notions of literal and variable forgetting) for nonclassical logics. This is what has been done (at least partly) in (Lakemeyer, 1997). Here we briefly consider two cases:

(i) *subclassical logics.* These logics are built on a classical language but have a weaker consequence relation than classical logic. For instance, in most multivalued logics, due to the fact that the excluded middle ($a \vee \neg a$) is not a theorem, a formula such as $a \vee (\neg b \wedge b)$ will depend both on $a$ and on $b$ (whereas it depends only on $a$ in classical logic), because it cannot be rewritten equivalently into a formula in which $b$ does not appear (in particular, $a$ is not always equivalent to $a \vee (\neg b \wedge b)$ in such logics); on the contrary, the formula $a \vee (a \wedge b)$ is equivalent to $a$ in usual multivalued logics and thus depends on $a$ only.

(ii) *modal logics.* Now, the language is obtained by extending a classical propositional language with one or more modalities, while the consequence relation extends that of classical logic (in the sense that a modality-free formula is a theorem if and only if it is a theorem of classical logic). Therefore, results such as Propositions 6 and 7 still make sense and we conjecture that they are still valid. To take an example, in the logic $S5$ where $\Box \Sigma \Rightarrow \Diamond \Sigma$ is a theorem, a formula such as $\Box \Sigma \wedge \Diamond (\Sigma \vee \Phi)$ is independent from $\Phi$ while $\Diamond \Sigma \wedge \Box (\Sigma \vee \Phi)$ is not.

### 6.2.3 Restricted queries and characteristic models

Forgetting a set of variables modifies a knowledge base while preserving some of the consequences, namely, those built on variables that are not forgotten. Other frameworks are based on restricting the possible queries in other ways, i.e., considering only queries in Horn form (Kautz, Kearns, & Selman, 1995).

---

6. This is not the case for single probability distributions: when the knowledge base is represented by a probability distribution $pr$, (a) does not lead to an acceptable definition (because a full probability distribution on $\Omega_{PS}$ always tell something about $x$); (c) suggests the definition $\forall \omega \in \Omega_{PS}, pr(Switch(\omega, x)) = pr(\omega)$, which leads to the decomposability of $pr$ w.r.t. $x$: $pr$ is a joint probability distribution obtained from a probability distribution $pr_{PS \setminus \{x\}}$ on $\Omega_{PS \setminus \{x\}}$ and the probability distribution $pr_x$ on $\{x\}$ defined by $pr_x(x) = \frac{1}{2}(= pr_x(\neg x))$. (b) would lead to the same definition, noticing that a full probability distribution can be expressed more compactly by a "partial" probability distribution (here, a probability distribution on $\Omega_{PS \setminus \{x\}}$) from which $pr$ is induced by the maximum entropy principle.

Reasoning with characteristic models (Kautz, Kearns, & Selman, 1993) is based on using only a subset of models of the original formula. As for forgetting, this may increase the size of the representation of a knowledge base exponentially. Khardon and Roth (1996) have shown that characteristic models can be used for efficient reasoning on some sets of restricted queries.

## 7. Concluding Remarks

We have investigated several ways of answering the key question of determining what a propositional knowledge base tells about the (in)dependencies between variables and formulas. For each of the notions studied, we have related it to other previously known notions and we have studied it from a computational point of view, giving both complexity results and characterization results to be used for practical computation. In the light of our results, it appears that the various forms of logical independence are closely connected. Especially, several of them had been proposed by different authors without being explicitely related. Boutilier's influenceability and Lakemeyer's relevance of a formula to a subject matter are identical to FV-independence (Propositions 25 and 26). We also discussed much related work, and suggest some extensions to more general framework than mere propositional logic.

The following table gives a synthetic view of many notions addressed in this paper and the corresponding complexity results.

| Problem | Notation | Definition | Complexity |
|---|---|---|---|
| Synt. Lit-independence | | $Lit(\Sigma) \cap L = \emptyset$ | P |
| Lit-independence | $L \not\rightarrow \Sigma$ | $\exists \Phi . \ \Phi \equiv \Sigma, \ Lit(\Phi) \cap L = \emptyset$ | coNP-complete |
| Dependent literals | $DepLit(\Sigma)$ | $\{l \mid \{l\} \mapsto \Sigma\}$ | $BH_2$-complete |
| FullLit-independence | | $L \subseteq DepLit(\Sigma)$ | coNP-complete |
| Lit-simplified | | $\forall l . \{l\} \not\rightarrow \Sigma \Leftrightarrow l \notin Lit(\Sigma)$ | NP-complete |
| Lit-equivalence | $\Sigma \equiv_L \Phi$ | $ForgetLit(\Sigma, Lit(\Sigma) \setminus L) \equiv$ | |
| | | $ForgetLit(\Phi, Lit(\Phi) \setminus L)$ | $\Pi_2^p$-complete |

Problems on literals and their complexity.

| Problem | Notation | Definition | Complexity |
|---|---|---|---|
| Synt. Var-independence | | $Var(\Sigma) \cap V = \emptyset$ | P |
| Var-independence | $V \not\rightarrow_-^+ \Sigma$ | $\exists \Phi . \ \Phi \equiv \Sigma, \ Var(\Phi) \cap \Sigma = \emptyset$ | coNP-complete |
| Dependent variables | $DepVar(\Sigma)$ | $\{v \mid v \mapsto_-^+ \Sigma\}$ | $BH_2$-complete |
| FullVar-independence | | $V \subseteq DepVar(\Sigma)$ | coNP-complete |
| Var-simplified | | $\forall v . \{v\} \not\rightarrow_-^+ \Sigma \Leftrightarrow v \notin Var(\Sigma)$ | NP-complete |
| Var-equivalence | $\Sigma \equiv_V \Phi$ | $ForgetVar(\Sigma, Var(\Sigma) \setminus V) \equiv$ | |
| | | $ForgetVar(\Phi, Var(\Phi) \setminus V)$ | $\Pi_2^p$-complete |

Problems on variables and their complexity.

The fact that both notions of formula-variable independence and forgetting have been used as key concepts in many AI fields (including automated reasoning, belief revision and update, diagnosis, reasoning about actions etc.) has been discussed before (Sections 1 and 4), so we will refrain from repeating it here. The gain of generality offered by the corresponding literal-oriented notions introduced in this paper has also been established (e.g., Proposition 20), and their application to several AI problems (like closed-world reasoning and belief update) has been sketched.

Primarily, one of the main motivations for the notions of formula-variable independence and forgetting was to improve inference from a computational side, by enabling to focus on relevant pieces of knowledge. The extent to which this goal can be reached can be discussed at the light of our complexity results:

- *Most (in)dependence relations have a high complexity.* The notions connected to FV-dependence (FL-dependence, full FV- and FL-dependence, influenceability, relevance to a subject matter and strict relevance second form) have a complexity at the first level of the polynomial hierarchy, which means that they can be checked by a satisfiability or/and an unsatisfiability solver. They become "tractable" when syntactical restrictions are made (Proposition 11). Forgetting (literals or variables) also is computationally expensive. The remaining notions are in complexity classes located at the second level of the polynomial hierarchy. Worse, under the standard assumptions of complexity theory, the explicit computation of literal or variable forgetting cannot be achieved in polynomial space in the general case (Proposition 23). This pushes towards the negative conclusion that all these notions are hard to be computed (at least in the worst case) except if the size or the syntactical form of the input enables it. The fact that these problems fall in the second level of the polynomial hierarchy are not that surprising since this is where a large part (if not the majority) of important problems in knowledge representation[7] fall.

- *But a high worst-case complexity does not necessarily prevent from practical algorithms!* Thus, Amir and McIlraith have shown the computational benefits that can be achieved by structuring a KB (through the forgetting operation) so as to achieve inference and consequence finding more efficiently (Amir & McIlraith, 2000; McIlraith & Amir, 2001). A similar discrepancy between the worst case situation and the practical ones can be observed in other domains; especially, satisfiability-based checkers for quantified boolean formulas (Biere, Cimatti, Clarke, Fujita, & Zhu, 1999; Williams, Biere, Clarke, & Gupta, 2000) used for formal verification purposes (bounded model checking) exhibit interesting computational behaviours (actually, they typically perform better than specialized algorithms, as shown in Rintanen, 2001), despite the fact that they are confronted to the problem of variable forgetting (i.e., elimination of existentially quantified variables).

- *Moreover, preprocessing may play an important role.* What we mean with "preprocessing" refers to the task of computing (in)dependence relations and forgetting *before* performing more problem-oriented tasks such as consequence finding, diagnosis, action/update, decision making etc. Thus, Lit-simplifying (or Var(simplifying) a KB during a preliminary off-line phase can prove helpful for improving on-line inference since simplification never increases the size of a KB. As shown by Proposition 23, a similar conclusion cannot be drawn to what concerns forgetting. This seems to be the price to be paid to benefit from the power of forgetting. However, this negative conclusion must be tempered by the two following comments. On the one hand, forgetting is interesting *per se*; it is not only a tool that can help improving inference in some cases but also a goal in several AI applications. On the other hand, our complexity results relate to the worst case situation, only, and, as evoked before, forgetting is feasible in many practical cases. Finally, let us note that there are several complete propositional fragments for which forgetting is easy. Especially, as advocated by Darwiche (1999), compiling a KB into a DNNF formula during an off-line step can prove practically valuable to achieve forgetting in an efficient way, provided that the size of the compiled form remains small enough (which cannot be guaranteed in the worst case). Since it is not known whether the DNNF fragment is strictly more succinct than the prime implicates one (Darwiche & Marquis, 1999), the prime implicates fragment can also be targeted with profit as a compilation language for some knowledge bases; especially, some recent approaches to the implicit representation of prime implicates (Simon & del Val, 2001) exhibit very significant empirical performances (they enable the computation of sets of prime implicates containing up to $10^{70}$ clauses). Accordingly, they can prove valuable for the practical computing of independence and forgetting.

---

7. Such as abduction, nonmonotonic inference, belief revision, belief update, some forms of planning and decision making.

## Acknowledgements

The third author has been partly supported by the IUT de Lens, the Université d'Artois, the Région Nord / Pas-de-Calais under the TACT-TIC project, and by the European Community FEDER Program. Some results of this paper already appeared in Section 3 of the paper (Lang & Marquis, 1998a) "Complexity results for independence and definability in propositional logic", *Proc. of the 6$^{th}$ International Conference on Principles of Knowledge Representation and Reasoning (KR'98)*, pages 356-367, 1998.

## Appendix A: Proofs

**Proposition 1**  *A formula $\Sigma$ is Lit-independent from $l$ if and only if, for any interpretation $\omega \in \Omega_{PS}$, if $\omega \models \Sigma$ then $Force(\omega, \neg l) \models \Sigma$.*

**Proof:** Assume that $\Sigma$ is Lit-independent from $l$. Then, there exists a formula $\Phi$ in NNF that is equivalent to $\Sigma$, and does not contain $l$. Then, for any $\omega \in \Omega_{PS}$ such that $\omega \models \Phi$ we have $Force(\omega, \neg l) \models \Phi$. Since $\Phi$ is equivalent to $\Sigma$, we conclude that the same property holds for $\Sigma$.

Assume that, for any interpretation $\omega \in \Omega_{PS}$, $\omega \models \Sigma$ implies $Force(\omega, \neg l) \models \Sigma$. We prove that $\Sigma$ is Lit-independent from $l$. Indeed, let $\gamma_\omega$ be the term whose only model is $\omega$. The following equivalence holds:

$$
\begin{aligned}
\Sigma &\equiv \bigvee\{\gamma_\omega \mid \omega \models \Sigma\} \\
&\equiv \bigvee\{\gamma_\omega \mid \omega \models \Sigma \text{ and } \omega \not\models l\} \cup \{\gamma_\omega \mid \omega \models \Sigma \text{ and } \omega \models l\} \\
&\equiv \bigvee\{\gamma_\omega \mid \omega \models \Sigma \text{ and } \omega \not\models l\} \cup \{\gamma_\omega \vee \gamma_{Force(\omega, \neg l)} \mid \omega \models \Sigma \text{ and } \omega \models l\}
\end{aligned}
$$

The latter step can be done because $\omega \models \Sigma$ implies that $Force(\omega, \neg l)$ is also a model of $\Sigma$. Now, if $\omega \not\models l$ then $\gamma_\omega$ does not contain $l$. On the other hand, if $\omega \models l$, then $\gamma_\omega \vee \gamma_{Force(\omega, \neg l)}$ can be rewritten as a conjunction of literals not containing neither $l$ nor its negation. As a result, the above formula (which is in NNF) does not contain $l$, which means that $\Sigma$ is Lit-independent from $l$. $\diamond$

**Proposition 2**
*(1) $DepLit(\Sigma) \subseteq Lit(\Sigma)$;*
*(2) If $\Sigma \equiv \Phi$, then $DepLit(\Sigma) = DepLit(\Phi)$;*
*(3a) $DepLit(\Sigma \wedge \Phi) \subseteq DepLit(\Sigma) \cup DepLit(\Phi)$;*
*(3b) $DepLit(\Sigma \vee \Phi) \subseteq DepLit(\Sigma) \cup DepLit(\Phi)$;*
*(4) $l \in DepLit(\Sigma)$ if and only if $\neg l \in DepLit(\neg\Sigma)$.*

**Proof:**

1. Trivial.

2. $L \mapsto \Sigma$ if and only if there exists a formula $\Psi$ that is equivalent to $\Sigma$, and such that $\Psi$ is syntactically Lit-independent from $L$. Since $\Sigma \equiv \Phi$, it follows that $\Phi \equiv \Psi$.

3. Let $\Psi$ (resp. $\Pi$) be a NNF formula equivalent to $\Sigma$ (resp. $\Phi$) s.t. no literal of $L$ occurs in it. Then $\Psi \wedge \Pi$ (resp. $\Psi \vee \Pi$) is a formula equivalent to $\Sigma \wedge \Phi$ (resp. $\Sigma \vee \Phi$), which is in NNF, and no literal of $L$ occurs in it.

4. Straightforward from the fact that $l$ appears in a NNF formula $\Sigma$ if and only if $\neg l$ appears in the NNF form of $\neg\Sigma$.

$\diamond$

**Proposition 3**
*(1) $DepVar(\Sigma) \subseteq Var(\Sigma)$;*
*(2) If $\Sigma \equiv \Phi$, then $DepVar(\Sigma) = DepVar(\Phi)$;*
*(3a) $DepVar(\Sigma \wedge \Phi) \subseteq DepLit(\Sigma) \cup DepLit(\Phi)$;*
*(3b) $DepVar(\Sigma \vee \Phi) \subseteq DepLit(\Sigma) \cup DepLit(\Phi)$;*
*(4) $DepVar(\neg\Sigma) = DepVar(\Sigma)$.*

**Proof:** (1) is trivial; (2) and (3) are similar to the proof of points (2) and (3) of Proposition 2, replacing "literal" by "variable", $l$ by $v$, and $DepLit$ by $DepVar$. As to (4): if $x \notin DepVar(\Sigma)$ then there exists a formula $\Phi$ equivalent to $\Sigma$ in which $x$ does not appear; since $x$ does not appear in $\neg\Phi$ either, $\neg\Phi$ is a formula equivalent to $\neg\Sigma$ in which $x$ does not appear. ◇

**Proposition 4** *Let $\Sigma$ be a formula from $PROP_{PS}$.*

- *$\Sigma$ is Lit-simplified if and only if the following equivalence holds: for every $L \subseteq L_{PS}$, $\Sigma$ is syntactically Lit-independent from $L$ if and only if $L \not\rightarrow \Sigma$ holds.*

- *$\Sigma$ is Var-simplified if and only if the following equivalence holds: for every $V \subseteq PS$, $\Sigma$ is syntactically Var-independent from $V$ if and only if $V \not\rightarrow^+_- \Sigma$ holds.*

**Proof:**

- Lit-simplification

    - ⇒: Assume $\Sigma$ is Lit-simplified (thus $Lit(\Sigma) = DepLit(\Sigma)$) and let $L \subseteq L_{PS}$. If $\Sigma$ is syntactically Lit-independent from $L$, then $L \cap Lit(\Sigma) = \emptyset$, thus $L \cap DepLit(\Sigma) = \emptyset$, i.e., $L \not\rightarrow \Sigma$.

    - ⇐: Assume that $\Sigma$ is not Lit-simplified. Then there exists $l \in Lit(\Sigma)$ s.t. $\Sigma$ is Lit-independent from $l$. With $L = \{l\}$, it is clear that $\Sigma$ is syntactically Lit-dependent on $L$, while Lit-independent from it, contradiction.

- Var-simplification. The proof is similar to the Lit-simplification case, replacing "Lit-independent" by "Var-independent", $L$ by $V$, $L_{PS}$ by $PS$, $l$ by $x$, $Lit(\Sigma)$ by $Var(\Sigma)$.

◇

**Proposition 5** *For every $\Sigma$ from $PROP_{PS}$, there exists a Lit-simplified formula $\Phi$ s.t. $\Sigma \equiv \Phi$.*

**Proof:** Since $\Sigma$ is Lit-independent from $L \setminus DepLit(\Sigma)$ we know that there exists a NNF formula $\Phi$ equivalent to $\Sigma$ such that $Lit(\Phi) \cap (L \setminus DepLit(\Sigma)) = \emptyset$, i.e., such that $Lit(\Phi) \subseteq DepLit(\Sigma)$. By point (2) of Proposition 2 we have $DepLit(\Phi) = DepLit(\Sigma)$. Thus, $DepLit(\Phi) \subseteq Lit(\Phi)) \subseteq DepLit(\Sigma) = DepLit(\Phi)$, from which we conclude that $DepLit(\Phi) = Lit(\Phi)$, i.e., $\Phi$ is Lit-simplified. ◇

**Proposition 6** *Let $\Sigma$ be a formula from $PROP_{PS}$ and $l$ be a literal of $L_{PS}$. The next four statements are equivalent:*

*(1)* $l \not\mapsto \Sigma$;
*(2)* $\Sigma_{l\leftarrow 1} \models \Sigma_{l\leftarrow 0}$;
*(3)* $\Sigma \models \Sigma_{l\leftarrow 0}$;
*(4)* $\Sigma_{l\leftarrow 1} \models \Sigma$.

**Proof:**

**(1) $\Rightarrow$ (2):** Let $v$ the variable of $l$ and assume that $\Sigma_{l\leftarrow 1} \not\models \Sigma_{l\leftarrow 0}$, which means that there is a $PS \setminus \{v\}$-world $\omega$ such that $\omega \models \Sigma_{l\leftarrow 1} \wedge \neg\Sigma_{l\leftarrow 0}$; since $\Sigma$ is equivalent to $(l \wedge \Sigma_{l\leftarrow 1}) \vee (\neg l \wedge \Sigma_{l\leftarrow 0})$, we have $\omega \wedge \{l\} \models \Sigma_{l\leftarrow 1} \wedge l \models \Sigma$ and $Force(\omega \wedge \{l\}, \neg l) = \omega \wedge \{\neg l\} \models \Sigma_{l\leftarrow 1} \wedge \neg\Sigma_{l\leftarrow 0}$; hence, $Force(\omega \wedge \{l\}, \neg l) \not\models \Sigma$ and therefore $l \mapsto \Sigma$ by Proposition 1.

**(2) $\Rightarrow$ (3):** Assume $\Sigma_{l\leftarrow 1} \models \Sigma_{l\leftarrow 0}$. We have the following chain of implications:

$$
\begin{aligned}
\Sigma &\equiv (l \wedge \Sigma_{l\leftarrow 1}) \vee (\neg l \wedge \Sigma_{l\leftarrow 0}) \\
\Sigma &\models (l \wedge \Sigma_{l\leftarrow 0}) \vee (\neg l \wedge \Sigma_{l\leftarrow 0}) \\
\Sigma &\models \Sigma_{l\leftarrow 0}
\end{aligned}
$$

**(3) $\Rightarrow$ (1):** Let us assume that $\Sigma \models \Sigma_{l\leftarrow 0}$, and prove that $l \not\mapsto \Sigma$. Indeed, the assumption can be rewritten as:

$$\forall \omega \in \Omega_{PS} \ . \ (\omega \models \Sigma \ \Rightarrow \omega \models \Sigma_{l\leftarrow 0})$$

Now, $\omega \models \Sigma_{l\leftarrow 0}$ is equivalent to say that changing the truth value of $l$ to false, $\omega$ is still a model of $\Sigma$. In formulas,

$$\forall \omega \in \Omega_{PS} \ . \ (\omega \models \Sigma \ \Rightarrow Force(\omega, \neg l) \models \Sigma)$$

This is exactly the definition of $l \not\mapsto \Sigma$.

**(2) $\Rightarrow$ (4):** Same as the proof of (2) $\Rightarrow$ (3).

**(4) $\Rightarrow$ (1):** Similar to the proof of (3) $\Rightarrow$ (1).

$\diamond$

**Proposition 7** *Let $\Sigma$ be a formula from $PROP_{PS}$ and $x$ be a variable of $PS$. The next four statements are equivalent:*
*(1) $x \not\mapsto^+_- \Sigma$;*
*(2) $\Sigma_{x\leftarrow 0} \equiv \Sigma_{x\leftarrow 1}$;*
*(3) $\Sigma \equiv \Sigma_{x\leftarrow 0}$;*
*(4) $\Sigma \equiv \Sigma_{x\leftarrow 1}$.*

**Proof**: Easy consequence of the definition of FV-independence ($\Sigma$ is Var-independent from $x$ if and only if it is Lit-independent from $\{x, \neg x\}$), and Proposition 6. $\diamond$

**Proposition 8** *Let $\Sigma$ be a formula from $PROP_{PS}$ and $L$ be a subset of $L_{PS}$. The next statements are equivalent:*

*(1) $L \not\rightarrow \Sigma$;*
*(2) $PI(\Sigma) \subseteq \{\gamma \mid \gamma$ is a term that does not contain any literal from $L\}$;*
*(3) $IP(\Sigma) \subseteq \{\delta \mid \delta$ is a clause that does not contain any literal from $L\}$.*

**Proof:**

- (1) $\Leftrightarrow$ (2):

  - $\Rightarrow$: If $\Sigma$ is Lit-independent from $L$, then there exists a NNF formula $\Phi$ equivalent to $\Sigma$ s.t. $Lit(\Phi) \cap L = \emptyset$. Clearly enough, the property $Lit(\Phi) \cap L = \emptyset$ is still satisfied if $\Phi$ is turned into DNF, especially when $\Phi$ is turned into its prime implicant normal form. Since two equivalent formulas have the same prime implicants, no term of $PI(\Sigma)$ contains a literal from $L$.

  - $\Leftarrow$: $PI(\Sigma)$ is a NNF formula that is equivalent to $\Sigma$ and syntactically Lit-independent from $L$. Hence, $\Sigma$ is Lit-independent from $L$.

- (1) $\Leftrightarrow$ (3): Similar to the prime implicant situation, using CNF instead of DNF.

$\diamond$

**Proposition 9** *Let $\Sigma$ be a formula from $PROP_{PS}$ and $V$ be a subset of $PS$. The next statements are equivalent:*
*(1) $V \not\rightarrow_{\perp}^{+} \Sigma$;*
*(2) $PI(\Sigma) \subseteq \{\gamma \mid \gamma$ is a term that does not contain any variable from $V\}$;*
*(3) $IP(\Sigma) \subseteq \{\delta \mid \delta$ is a clause that does not contain any variable from $V\}$.*

**Proof:** Easy consequence of the definition of FV-independence, plus Proposition 9 above. $\diamond$

**Proposition 10** FL DEPENDENCE, FV DEPENDENCE, FULL FL DEPENDENCE *and* FULL FV DEPENDENCE *are* NP-*complete.*

**Proof:**

- **FL dependence**

  - Membership. In order to show that $\Sigma$ is Lit-dependent on $L$, it is sufficient to guess a literal $l$ from $L$ and a $(Var(\Sigma) \setminus \{Var(l)\})$-world $\omega$ that is a model of $\Sigma_{l \leftarrow 1}$ but not a model of $\Sigma_{l \leftarrow 0}$. These tests can be achieved in time polynomial in $|\Sigma|$.

  - Hardness. Let us consider the mapping $M$ s.t. $M(\Sigma) = \langle \Sigma \wedge new, new \rangle$, where $new$ is a propositional variable that does not occur in $\Sigma$. Clearly enough, $M(\Sigma)$ can be computed in time polynomial in $|\Sigma|$. Moreover, $\Sigma$ is satisfiable if and only if $\Sigma \wedge new$ is Lit-dependent on $\{new\}$.

- **FV dependence**

- Membership. In order to show that $\Sigma$ is Var-dependent on $V$, it is sufficient to guess a variable $x$ from $V$ and a $(Var(\Sigma) \setminus \{x\})$-world $\omega$ that is not a model of the formula $\Sigma_{x \leftarrow 0} \Leftrightarrow \Sigma_{x \rightarrow 1}$. This test can be achieved in time polynomial in $|\Sigma|$.

- Hardness. Similar to the hardness proof of FL-dependence, replacing "Lit-dependent" by "Var-dependent".

- **full FL dependence**

  - Membership. $\Sigma$ is fully Lit-dependent on $L = \{l_1, ..., l_n\}$ if and only if $\Sigma_{l_1 \leftarrow 1} \not\models \Sigma_{l_1 \leftarrow 0}$ and ... and $\Sigma_{l_n \leftarrow 1} \not\models \Sigma_{l_n \leftarrow 0}$ holds. Equivalently, $\Sigma$ is fully Lit-dependent on $L = \{l_1, ..., l_n\}$ if and only if the formula $rename_1(\Sigma_{l_1 \leftarrow 1} \wedge \neg \Sigma_{l_1 \leftarrow 0}) \wedge ... \wedge rename_n(\Sigma_{l_n \leftarrow 1} \wedge \neg \Sigma_{l_n \leftarrow 0})$ is satisfiable, where each $rename_i$ is a renaming that maps variables to new symbols (in a uniform way). Since this formula can be computed in time polynomial in $|\Sigma| + |V|$, the membership of FULL FL DEPENDENCE to NP follows.

  - Hardness. Full FL-dependence and FL-dependence coincide in the case in which $L$ is composed of a single literal. Since the NP-hardness of FL-dependence has been proved using a set $L$ composed of a single literal, the NP-hardness of full FL dependence follows.

- **full FV dependence**

  - Membership. $\Sigma$ is fully Var-dependent on $V = \{x_1, ..., x_n\}$ if and only if $\Sigma \not\equiv \Sigma_{x_1 \leftarrow 0}$ and ... and $\Sigma \not\equiv \Sigma_{x_n \leftarrow 0}$ holds. Equivalently, $\Sigma$ is fully Var-dependent on $V = \{x_1, ..., x_n\}$ if and only if the formula $rename_1(\Sigma \oplus \Sigma_{x_1 \leftarrow 0}) \wedge ... \wedge rename_n(\Sigma \oplus \Sigma_{x_n \leftarrow 0})$ is satisfiable, where each $rename_i$ is a renaming that maps variables to new symbols (in a uniform way). Since this formula can be computed in time polynomial in $|\Sigma| + |V|$, the membership of FULL FV DEPENDENCE to NP follows.

  - Hardness. Full FV-dependence and FV-dependence coincide if $L$ is composed of a single literal. Since the NP-hardness of FV-dependence has been proved using a set $L$ composed of a single literal, the NP-hardness of full FV-dependence follows.

$\diamond$

**Proposition 11** *Whenever $\Sigma$ belongs to a class $\mathcal{C}$ of CNF formulas that is tractable for clausal query answering (i.e., there exists a polytime algorithm to determine whether $\Sigma \models \gamma$ for any CNF formula $\gamma$) and stable for variable instantiation (i.e., replacing in $\Sigma \in \mathcal{C}$ any variable by* true *or by* false *gives a formula that still belongs to $\mathcal{C}$) then* FL DEPENDENCE, FV DEPENDENCE, FULL FL DEPENDENCE *and* FULL FV DEPENDENCE *are in* P.

**Proof:** This is a straightforward consequence of Propositions 6 and 7. When $\Sigma$ belongs to a class $\mathcal{C}$ of formulas that is tractable for clausal query answering and stable for variable instantiation, we can easily check whether $\Sigma \models \Sigma_{x \leftarrow 0}$ and $\Sigma_{x \leftarrow 0} \models \Sigma$ holds. $\diamond$

**Proposition 12** Lit-simplified formula *and* Var-simplified formula *are* NP-*complete*.

**Proof:**

- **Lit-simplification**

  - Membership. Easy consequence of the fact that Lit-simplified formula is a restriction of full FL dependence that is in NP (and NP is closed under polynomial reductions).

  - Hardness. We prove that a formula $\Phi$, built over an alphabet $X = \{x_1, \ldots, x_n\}$, is satisfiable if and only if formula $\Sigma$ is L-simplified, where

  $$\Sigma = (\Phi \vee \Phi[X/\neg X]) \wedge \bigwedge_{x_i \in X} x_i \Leftrightarrow y_i$$

  where $\Phi[X/\neg X]$ is the formula obtained by replacing in $\Phi$ each occurrence of $x_i$ (resp. $\neg x_i$) with $\neg x_i$ (resp. $x_i$).

  First, if $\Phi$ is not satisfiable, neither $\Phi[X/\neg X]$ is, thus $(\Phi \vee \Phi[X/\neg X])$ is not satisfiable. As a result, $\Sigma$ is not satisfiable, thus it is not Lit-simplified, because $DepLit(\Sigma) = \emptyset$ but $\Sigma$ mentions variables $x_i$ and $y_i$.

  Assume $\Phi$ satisfiable. Clearly, $\Sigma$ is satisfiable as well: let $\omega$ be a model of $\Sigma$. We prove that $\Sigma$ is Lit-simplified by showing that it is Lit-dependent on each literal it contains. We have $Lit(\Sigma) = \{x_1, \ldots, x_n, y_1, \ldots, y_n, \neg x_1, \ldots, \neg x_n, \neg y_1, \ldots, \neg y_n\}$. Let $l_i \in Lit(\Sigma)$.

    (1) Assume that $\omega \models l_i$. Then, $Force(\omega, \neg l_i) \not\models \Sigma$. Indeed, let $Var(l_i) = x_i$ (resp. $y_i$): $\omega$ satisfies $x_i \Leftrightarrow y_i$ so changing the truth value of $x_i$ only (resp. $y_i$ only) leads to a model that does not satisfy $x_i \Leftrightarrow y_i$.

    (2) Otherwise, we have $\omega \models \neg l_i$. Replacing $l_i$ by $\neg l_i$ in the proof of case (1) enables deriving the expected conclusion.

  - **Var-simplification**

    * Membership. Easy consequence of the fact that Var-simplified formula is a restriction of full FV dependence that is in NP (and NP is closed under polynomial reduction).

    * Hardness. The proof is similar to the proof of NP-hardness of Lit-simplified formula, replacing "Lit-simplified" with "Var-simplified".

  $\diamond$

**Proposition 13**

1. *Determining whether $DepLit(\Sigma) = L$ (where $L$ is a set of literals), and determining whether $DepVar(\Sigma) = X$ (where $X$ is a set of variables) is* BH$_2$-*complete*.

2. *The search problem consisting in computing $DepLit(\Sigma)$ (respectively $DepVar(\Sigma)$) is in* F$\Delta_2^p$ *and is both* NP *and* coNP-*hard*.

**Proof**:

- Determining whether $DepLit(\Sigma) = L$ is $\mathsf{BH}_2$-complete.

    - Membership.

      $DepLit(\Sigma) = L$ if and only if (i) $l \mapsto \Sigma$ holds for every $l \in L$ and (ii) $l \not\mapsto \Sigma$ holds for every $l \in Lit(\Sigma) \setminus L$; the set of instances $\langle \Sigma, L \rangle$ such that (i) holds is the union of a linear number of problems in $\mathsf{NP}$, thus it is in $\mathsf{NP}$; similarly, the set of instances $\langle \Sigma, L \rangle$ such that (ii) holds is a problem in $\mathsf{coNP}$. This proves the membership to $\mathsf{BH}_2$ of the problem of determining whether $DepLit(\Sigma) = L$.

    - Hardness.

      Let $\langle \Phi, \Gamma \rangle$ be a pair of propositional formulas. Without loss of generality, we assume that $\Phi$ and $\Gamma$ do not share any variable and that $X = Var(\Phi) = \{x_1, ..., x_n\}$, $Var(\Gamma) = \{z_1, ..., z_p\}$. Furthermore we assume that $X \neq \emptyset$ (if it were not the case it would then be sufficient then to replace $\Phi$ by $\Phi \wedge (t \vee \neg t)$ before performing the reduction). We prove that $\langle \Phi, \Gamma \rangle$ is a positive instance of SAT-UNSAT, i.e., $\Phi$ is satisfiable and $\Gamma$ is unsatisfiable, if and only if $DepLit(\Phi \wedge \neg\Gamma) = Lit(\Sigma(\Phi))$ where $\Sigma(\Phi) = (\Phi \vee \Phi[X/\neg X]) \wedge \bigwedge_{x_i \in X} x_i \Leftrightarrow y_i$ as in a previous proof (note that $\Sigma(\Phi)$ and $\Gamma$ do not share any variables).

      (i) if $\Phi$ is satisfiable and $\Gamma$ is unsatisfiable, then $\Sigma(\Phi)$ is Lit-simplified according to the previous proof and thus $DepLit(\Sigma(\Phi)) = Lit(\Sigma(\Phi))$ whereas $DepLit(\neg\Gamma) = \emptyset$, which together entail that $DepLit(\Sigma(\Phi) \wedge \neg\Gamma) = Lit(\Sigma(\Phi))$.

      (ii) Assume that $\Phi$ is unsatisfiable and $DepLit(\Sigma(\Phi) \wedge \neg\Gamma) = Lit(\Sigma(\Phi))$. Then $\Sigma(\Phi)$ is not simplified (recall that $Var(\Phi)$ is not empty and hence so is $Lit(\Sigma(\Phi))$), thus $DepLit(\Sigma(\Phi)) \subset Lit(\Sigma(\Phi))$. Now, $DepLit(\Sigma(\Phi) \wedge \neg\Gamma) = Lit(\Sigma(\Phi))$ implies $DepLit(\Sigma(\Phi) \wedge \neg\Gamma) \cap Lit(\Gamma) = \emptyset$, which is possible only if $\Gamma$ is a tautology (because $\Sigma(\Phi)$ and $\Gamma$ do not share any variables); but in this case, $\Sigma(\Phi) \wedge \neg\Gamma$ is inconsistent and thus $DepLit(\Sigma(\Phi) \wedge \neg\Gamma) = \emptyset$, hence, $Lit(\Sigma(\Phi)) = \emptyset$, which is contradictory.

      (iii) Assume that $\Gamma$ is satisfiable and $DepLit(\Sigma(\Phi) \wedge \neg\Gamma) = Lit(\Sigma(\Phi))$.

      The second condition can hold only if $\Sigma(\Phi)$ is unsatisfiable, hence not Lit-simplified (since $Lit(\Sigma(\Phi)) \neq \emptyset$), so $\Phi$ is unsatisfiable as well; this takes us back to the case (ii), which leads us to the same contradiction again.

- Computing $DepLit(\Sigma)$ is in $\mathsf{F}\Delta_2^p$ since FL DEPENDENCE is in $\mathsf{NP}$ and $DepLit(\Sigma) \subseteq Lit(\Sigma)$ (hence, it is sufficient to test for every $l \in Lit(\Sigma)$ whether or not $\Sigma$ is Lit-independent from it). It is also both $\mathsf{NP}$-hard and $\mathsf{coNP}$-hard. $\mathsf{NP}$-hardness is showed by the following polynomial reduction from SAT: A CNF formula $F$ is satisfiable if and only if $F$ is valid or $DepLit(F) \neq \emptyset$; clearly enough, valid CNF formulas can be recognized in polynomial time. Thus, SAT can be solved if we know how to compute $DepLit(\Sigma)$ for any $\Sigma$, which shows that computing $DepLit(\Sigma)$ is $\mathsf{NP}$-hard. The proof of $\mathsf{coNP}$-hardness is similar ($F$ is unsatisfiable if and only if $F$ is not valid and $DepLit(F) = \emptyset$).

- Determining whether $DepVar(\Sigma) = V$ is $\mathsf{BH}_2$-complete. Membership follows easily from the membership of the corresponding problem for $DepLit(\Sigma)$. Hardness is similar to the hardness proof for $DepLit(\Sigma) = L$.

- Computing $DepVar(\Sigma)$ is in $\mathsf{F}\Delta_2^p$ and is both $\mathsf{NP}$-hard and $\mathsf{coNP}$-hard. Similar to the corresponding result for $DepLit(\Sigma)$.

◇

**Proposition 14** *The set of models of $ForgetLit(\Sigma, \{l\})$ can be expressed as:*

$$
\begin{aligned}
Mod(ForgetLit(\Sigma, \{l\})) &= Mod(\Sigma) \cup \{Force(\omega, \neg l) \mid \omega \models \Sigma\} \\
&= \{\omega \mid Force(\omega, l) \models \Sigma\}
\end{aligned}
$$

**Proof:** The proof is obtained immediately from the definition. ◇

**Proposition 15** *The set of models of $ForgetLit(\Sigma, L)$ can be expressed as:*

$$
Mod(ForgetLit(\Sigma, L)) = \{\omega \mid Force(\omega, L_1) \models \Sigma \text{ where } L_1 \subseteq L\}
$$

**Proof:** By induction on $|L|$. The base case in which $L$ is empty is trivial. Let now assume that the property holds for any $L$ composed of $k$ elements, and prove that it holds as well for $L \cup \{l\}$.

By definition,

$\omega \models ForgetLit(\Sigma, L \cup \{l\})$

    if and only if   $\omega \models ForgetLit(ForgetLit(\Sigma, L), \{l\})$

    if and only if   $\omega \models ForgetLit(\Sigma, L)$ or $Force(\omega, l) \models ForgetLit(\Sigma, L)$

    if and only if   $Force(\omega, L') \models ForgetLit(\Sigma, L)$ where $L' \subseteq \{l\}$

Using the induction hypothesis, we can express the set of models of $ForgetLit(\Sigma, L)$ as the models $\omega'$ such that $Force(\omega', L_1) \models \Sigma$, where $L_1$ is any subset of $L$. As a result,

$\omega \models ForgetLit(\Sigma, L \cup \{l\})$

    if and only if   $Force(\omega, L') = \omega'$ and $Force(\omega', L_1) \models \Sigma$, where $L' \subseteq \{l\}$ and $L_1 \subseteq L$

    if and only if   $Force(\omega, L_1') \models \Sigma$ where $L_1' \subseteq L \cup \{l\}$

As a result, the models of $ForgetLit(\Sigma, L)$ are the models that can be mapped into models of $\Sigma$ by forcing a subset of literals in $L$ to become true. ◇

**Proposition 16** *Let $\Sigma$ be a formula from $PROP_{PS}$ and $L \subseteq L_{PS}$. $ForgetLit(\Sigma, L)$ is the logically strongest consequence of $\Sigma$ that is Lit-independent from $L$ (up to logical equivalence).*

**Proof:** By induction on $|L|$. The base case $|L| = 0$ is trivial. Let us now assume that the proposition holds for every $|L| \leq k$ and show that it remains true when $|L| = k + 1$.

Let $L = L' \cup \{l\}$. By the induction hypothesis, we can assume that $ForgetLit(\Sigma, L')$ is the most general consequence of $\Sigma$ that is Lit-independent from $L'$. For the sake of simplicity, let $\Sigma'$ denote this formula. It remains to show that $ForgetLit(\Sigma, L) = ForgetLit(ForgetLit(\Sigma, L'), \{l\}) = ForgetLit(\Sigma', \{l\})$ is the most general consequence of $\Sigma'$ that is Lit-independent from $l$. Two cases are to be considered:

- $l = x$. $ForgetLit(\Sigma', x)$ is Lit-independent from $x$ if and only if $ForgetLit(\Sigma', x) \models ForgetLit(\Sigma', x)_{x \leftarrow 0}$ holds. We have

$$(\Sigma'_{x \leftarrow 1} \vee (\neg x \wedge \Sigma'))_{x \leftarrow 0} \equiv (\Sigma'_{x \leftarrow 1} \vee \Sigma'_{x \leftarrow 0})$$

  This formula clearly is a logical consequence of $ForgetLit(\Sigma', \{x\})$. Hence $ForgetLit(\Sigma', x)$ is Lit-independent from $x$. $ForgetLit(\Sigma', x)$ is a logical consequence of $\Sigma'$. Indeed, for every $\Sigma' \in PROP_{PS}$, $x \in PS$, we have $\Sigma' \equiv (x \wedge \Sigma'_{x \leftarrow 1}) \vee (\neg x \wedge \Sigma'_{x \leftarrow 0})$. It remains to show that every logical consequence $\Phi$ of $\Sigma'$ that is Lit-independent from $x$ is a logical consequence of $ForgetLit(\Sigma', x)$. Let $\Phi$ any formula s.t. $\Sigma' \models \Phi$ holds and $\Phi \models \Phi_{x \leftarrow 0}$. Since $\Sigma' \equiv (x \wedge \Sigma'_{x \leftarrow 1}) \vee (\neg x \wedge \Sigma'_{x \leftarrow 0})$ holds, we have $\Sigma' \models \Phi$ if and only if both $(x \wedge \Sigma'_{x \leftarrow 1}) \models \Phi$ and $(\neg x \wedge \Sigma'_{x \leftarrow 0}) \models \Phi$ hold. Thus, in order to show that $ForgetLit(\Sigma', x) \models \Phi$ holds, it is sufficient to prove that $\Sigma'_{x \leftarrow 1} \models \Phi$ holds. Suppose that is not the case. Then, there exists an implicant $\gamma$ of $\Sigma'_{x \leftarrow 1}$ that is not an implicant of $\Phi$. However, since $(x \wedge \Sigma'_{x \leftarrow 1}) \models \Phi$ holds, we know that $(x \wedge \gamma)$ is an implicant of $\Phi$. Since $\Phi$ is equivalent to the conjunction of its prime implicates, there necessarily exists a prime implicate $\pi$ of $\Phi$ s.t. $(x \wedge \gamma) \models \pi$ and $\gamma \not\models \pi$ hold. This imposes that $x$ belongs to $\pi$ but no literal of $\gamma$ belongs to $\pi$. By construction, $\pi_{x \leftarrow 0}$ is an implicate of $\Phi_{x \leftarrow 0}$ and $\pi_{x \leftarrow 0}$ is strictly stronger than $\pi$. Since $\Phi \models \Phi_{x \leftarrow 0}$ holds, $\Phi \models \pi_{x \leftarrow 0}$ holds as well. This contradicts the fact that $\pi$ is a prime implicate of $\Phi$.

- $l = \neg x$. The demonstration is similar, *mutatis mutandis*.

$\diamond$

**Proposition 17** *Let $\Sigma$, $\Phi$ be two formulas from $PROP_{PS}$ and $L \subseteq L_{PS}$.*

$$ForgetLit(\Sigma \vee \Phi, L) \equiv ForgetLit(\Sigma, L) \vee ForgetLit(\Phi, L).$$

**Proof:** The claim can be easily proved from Proposition 15. Indeed, the models of $ForgetLit(\Sigma \vee \Phi, L)$ are the models $\omega$ such that $Force(\omega, L_1) \models \Sigma \vee \Phi$, where $L_1 \subseteq L$. Now, $Force(\omega, L_1) \models \Sigma \vee \Phi$ holds if and only if $Force(\omega, L_1) \models \Sigma$ holds or $Force(\omega, L_1) \models \Phi$ holds.

On the other hand, the models of $ForgetLit(\Sigma, L) \vee ForgetLit(\Phi, L)$ are those such that $Force(\omega, L_1) \models \Sigma$ or $Force(\omega, L_1) \models \Phi$ for some $L_1 \subseteq L$. This is equivalent to the above condition.

$\diamond$

**Proposition 18** *Let $\gamma$ be a consistent term from $PROP_{PS}$ and $L \subseteq L_{PS}$.*
$ForgetLit(\gamma, L) \equiv \bigwedge_{l \in \gamma \, s.t. \, l \notin L} l.$

**Proof**: We first prove the following lemma:

**Lemma 1** *For any consistent term $\gamma$, we have $ForgetLit(\gamma, l) \equiv \gamma \setminus \{l\}$.*

    *Proof (lemma):*

**case 1** $l \in \gamma$, *i.e.,* $\gamma = l \wedge \gamma'$.
    $ForgetLit(\gamma, l) \equiv \gamma_{l \leftarrow 1} \vee (\neg l \vee \gamma) \equiv \gamma' \equiv \gamma \setminus \{l\}$.

**case 2** $\neg l \in \gamma$, *i.e.,* $\gamma = \neg l \wedge \gamma'$.
    $ForgetLit(\gamma, l) \equiv \bot \vee \gamma \equiv \gamma \equiv \gamma \setminus \{l\}$.

**case 3** $l \notin \gamma$ and $\neg l \notin \gamma$.
    $ForgetLit(\gamma, l) \equiv \gamma \vee (\neg l \vee \gamma) \equiv \gamma \equiv \gamma \setminus \{l\}$.

                                                                                              $\diamond$

    A straightforward induction on $L$ completes the proof.               $\diamond$

**Proposition 19** *Let $\Sigma$ be a formula from $PROP_{PS}$ and $L \subseteq L_{PS}$.*
$IP(ForgetLit(\Sigma, L)) = \{\delta \mid \delta \in IP(\Sigma) \text{ and } Lit(\delta) \cap L = \emptyset\}$.

**Proof:**

- $\subseteq$: Let $\delta \in IP(ForgetLit(\Sigma, L))$. Since $\Sigma \models ForgetLit(\Sigma, L)$ and $ForgetLit(\Sigma, L) \models \delta$ hold, $\delta$ is an implicate of $\Sigma$. Hence, there exists a prime implicate $\delta'$ of $\Sigma$ s.t. $\delta' \models \delta$ holds. Since $\delta$ does not contain any literal from $L$, this is also the case for $\delta'$. Thus, $\delta'$ is a logical consequence of $\Sigma$ that is Lit-independent from $L$. As a consequence of Proposition 16, it must be the case that $ForgetLit(\Sigma, L) \models \delta'$. Hence, there exists a prime implicate $\delta''$ of $ForgetLit(\Sigma, L)$ s.t. $\delta'' \models \delta'$ holds. This implies that $\delta'' \models \delta$ holds as well, and since both clauses are prime implicates of the same formula, they are equivalent. But this implies that $\delta \equiv \delta'$ holds, which completes the proof.

- $\supseteq$: Let $\delta$ be a prime implicate of $\Sigma$ that does not contain any literal from $L$. $\delta$ is Lit-independent from $L$. Since $\delta$ is an implicate of $\Sigma$, it must also be an implicate of $ForgetLit(\Sigma, L)$. Subsequently, there exists a prime implicate $\delta'$ of $ForgetLit(\Sigma, L)$ s.t. $\delta' \models \delta$ holds. Since $\Sigma \models ForgetLit(\Sigma, L)$ and $ForgetLit(\Sigma, L) \models \delta'$ both hold, we have $\Sigma \models \delta'$ as well. Hence, there exists a prime implicate $\delta''$ of $\Sigma$ s.t. $\delta'' \models \delta'$ holds. Therefore, $\delta'' \models \delta$ holds and since both clauses are prime implicates of the same formula, they are equivalent. But this implies that $\delta \equiv \delta'$ holds, which completes the proof.

                                                                                            $\diamond$

**Proposition 20** *Let $\Sigma$ be a formula from $PROP_{PS}$ and $V \subseteq PS$. We have*

$$ForgetVar(\Sigma, V) \equiv ForgetLit(\Sigma, L_V)$$

**Proof:** By induction on $|V|$. The proposition trivially holds for $|V| = 0$. Let us assume it is true whenever $|V| = k$. Let $V = \{v_1, ..., v_{k+1}\}$. By definition, we have

$$ForgetVar(\Sigma, V) = ForgetVar(ForgetVar(\Sigma, \{v_1, ..., v_k\}), \{v_{k+1}\})$$

By the induction hypothesis, $ForgetVar(\Sigma, V)$ is equivalent to $ForgetVar(\Phi, \{v_{k+1}\})$, where $\Phi$ is defined as:

$$\Phi = ForgetLit(\Sigma, \{x; x \in \{v_1, \ldots, v_k\}\} \cup \{\neg x; x \in \{v_1, \ldots, v_k\}\})$$

We have $ForgetVar(\Sigma, V) \equiv ForgetVar(\Phi, \{v_{k+1}\})$. By definition,

$$ForgetVar(\Phi, v_{k+1}) = \Phi_{v_{k+1} \leftarrow 0} \vee \Phi_{v_{k+1} \leftarrow 1}.$$

We also have

$ForgetLit(\Phi, \{v_{k+1}, \neg v_{k+1}\})$
  $= \; ForgetLit(ForgetLit(\Phi, \{v_{k+1}\}), \{\neg v_{k+1}\})$
  $= \; ForgetLit((\neg v_{k+1} \wedge \Phi_{v_{k+1} \leftarrow 0}) \vee \Phi_{v_{k+1} \leftarrow 1}, \{\neg v_{k+1}\})$
  $= \; (v_{k+1} \wedge ((\neg v_{k+1} \wedge \Phi_{v_{k+1} \leftarrow 0}) \vee \Phi_{v_{k+1} \leftarrow 1})_{v_{k+1} \leftarrow 1}) \vee ((\neg v_{k+1} \wedge \Phi_{v_{k+1} \leftarrow 0}) \vee \Phi_{v_{k+1} \leftarrow 1})_{v_{k+1} \leftarrow 0}.$

This simplifies to $(v_{k+1} \wedge \Phi_{v_{k+1} \leftarrow 1}) \vee \Phi_{v_{k+1} \leftarrow 0} \vee \Phi_{v_{k+1} \leftarrow 1}$, which is also equivalent to $\Phi_{v_{k+1} \leftarrow 0} \vee \Phi_{v_{k+1} \leftarrow 1}$, hence equivalent to $ForgetVar(\Phi, \{v_{k+1}\})$. Consequently, we have

$ForgetVar(\Sigma, V)$
  $\equiv \; ForgetLit(ForgetLit(\Sigma, \{x; x \in \{v_1, \ldots, v_k\}\} \cup \{\neg x; x \in \{v_1, \ldots, v_k\}\}), \{v_{k+1}, \neg v_{k+1}\})$
  $\equiv \; ForgetLit(\Sigma, V).$

$\diamond$

**Proposition 21** *Let $\Sigma$, $\Phi$ be two formulas from $PROP_{PS}$, and $V$ be a subset of $PS$. If $V \not\rightarrow^+_- \Sigma$, then $ForgetVar(\Sigma \wedge \Phi, V) \equiv \Sigma \wedge ForgetVar(\Phi, V)$.*

**Proof:** Let us consider the case where $V = \{v\}$. By definition, we have $ForgetVar(\Sigma \wedge \Phi, \{v\}) = (\Sigma \wedge \Phi)_{v \leftarrow 0} \vee (\Sigma \wedge \Phi)_{v \leftarrow 1}$. Equivalently, $ForgetVar(\Sigma \wedge \Phi, \{v\}) \equiv (\Sigma_{v \leftarrow 0} \wedge \Phi_{v \leftarrow 0}) \vee (\Sigma_{v \leftarrow 1} \wedge \Phi_{v \leftarrow 1})$. When $v \not\rightarrow^+_- \Sigma$, we have $\Sigma \equiv \Sigma_{v \leftarrow 0} \equiv \Sigma_{v \leftarrow 1}$. Accordingly, $ForgetVar(\Sigma \wedge \Phi, \{v\}) \equiv \Sigma \wedge (\Phi_{v \leftarrow 0} \vee \Phi_{v \leftarrow 1}) \equiv \Sigma \wedge Forget(\Phi, \{v\})$. A straightforward induction completes the proof. $\diamond$

**Proposition 22** *Let $\Sigma$, $\Phi$ be two formulas from $PROP_{PS}$, and $P$, $Q$, $Z$ be three disjoint sets of variables from $PS$ (such that $Var(\Sigma) \cup Var(\Phi) \subseteq P \cup Q \cup Z$). It holds:*

1. *If $\Phi$ does not contain any variable from $Z$*

$$CIRC(\Sigma, \langle P, Q, Z \rangle) \models \Phi$$
$$\textit{if and only if}$$
$$\Sigma \models ForgetLit(\Sigma \wedge \Phi, L_P^- \cup L_Z)$$

2. *In the general case:*

$$CIRC(\Sigma, \langle P, Q, Z \rangle) \models \Phi$$
$$\textit{if and only if}$$
$$\Sigma \models ForgetLit(\Sigma \wedge \neg ForgetLit(\Sigma \wedge \neg \Phi, L_Z \cup L_P^-), L_Z \cup L_P^-)$$

*where $CIRC$ is circumscription as defined in (McCarthy, 1986).*

**Proof:**

1. This is a consequence of Theorem 2.5 from (Przymusinski, 1989). This theorem states that if $\Phi$ does not contain literals from $Z$, then $CIRC(\Sigma, \langle P, Q, Z \rangle) \models \Phi$ holds if and only if there is no clause $\gamma$ s.t. $\gamma$ does not contain any literal from $L_P^- \cup L_Z$ and $\Sigma \models \neg \Phi \vee \gamma$ but $\Sigma \not\models \gamma$. This is equivalent to state that, if $\Phi$ does not contain literals from $Z$, then $CIRC(\Sigma, \langle P, Q, Z \rangle) \models \Phi$ holds if and only if, for every clause $\gamma$ containing only literals from $L_P^+ \cup L_Q$, we have $\Sigma \wedge \Phi \models \gamma$ if and only if $\Sigma \models \gamma$. It is easy to see that the equivalence is preserved would any formula $\gamma$ Lit-independent from $L_P^- \cup L_Z$ be considered (any formula can be turned into an equivalent CNF formula). Thus, Theorem 2.5 can be rephrased in forgetting terms: if $\Phi$ does not contain variables from $Z$, then $CIRC(\Sigma, \langle P, Q, Z \rangle) \models \Phi$ holds if and only if $\Sigma \wedge \Phi \equiv_{L_P^+ \cup L_Q} \Sigma$ if and only if $\Sigma \models ForgetLit(\Sigma \wedge \Phi, L_P^- \cup L_Z)$.

2. This is a consequence of Theorem 2.6 from (Przymusinski, 1989). This theorem states that $CIRC(\Sigma, \langle P, Q, Z \rangle) \models \Phi$ holds if and only if $\Sigma \models \Phi$ or there exists a formula $\Psi$ s.t. $\Psi$ does not contain any literal from $L_P^- \cup L_Z$, $\Sigma \models \Phi \vee \Psi$ holds and $CIRC(\Sigma, \langle P, Q, Z \rangle) \models \neg \Psi$. It is easy to see that such a formula $\Psi$ exists if and only if the conjunction $\alpha$ of all the formulas $\Psi$ such that $\Psi$ does not contain any literal from $L_P^- \cup L_Z$ and $\Sigma \models \Phi \vee \Psi$ holds is s.t. $CIRC(\Sigma, \langle P, Q, Z \rangle) \models \neg \alpha$. Since $\Sigma \models \Phi \vee \Psi$ holds if and only if $\Sigma \wedge \neg \Phi \models \Psi$ holds, $\alpha$ is equivalent to $ForgetLit(\Sigma \wedge \neg \Phi, L_P^- \cup L_Z)$. Thus, $CIRC(\Sigma, \langle P, Q, Z \rangle) \models \Phi$ holds if and only if $\Sigma \models \Phi$ holds or $CIRC(\Sigma, \langle P, Q, Z \rangle) \models \neg ForgetLit(\Sigma \wedge \neg \Phi, L_P^- \cup L_Z)$ holds. In the case where $\Sigma \models \Phi$ holds, $\Sigma \wedge \neg \Phi$ is inconsistent, so it is also the case of $ForgetLit(\Sigma \wedge \neg \Phi, L_P^- \cup L_Z)$. Thus, if $\Sigma \models \Phi$ holds, $CIRC(\Sigma, \langle P, Q, Z \rangle) \models \neg ForgetLit(\Sigma \wedge \neg \Phi, L_P^- \cup L_Z)$ holds as well. Accordingly, $CIRC(\Sigma, \langle P, Q, Z \rangle) \models \Phi$ holds if and only if $CIRC(\Sigma, \langle P, Q, Z \rangle) \models \neg ForgetLit(\Sigma \wedge \neg \Phi, L_P^- \cup L_Z)$ holds. Since $ForgetLit(\Sigma \wedge \neg \Phi, L_P^- \cup L_Z)$ does not contain any literal from $Z$, the point just above enables concluding the proof.

◇

**Proposition 23** *Let $\Sigma$ be a formula from $PROP_{PS}$ and let $L$ be a subset of $L_{PS}$. In the general case, there is no propositional formula $\Phi$ equivalent to $ForgetLit(\Sigma, L)$ s.t. the size of $\Phi$ is polynomially bounded in $|\Sigma| + |L|$, unless $\mathsf{NP} \cap \mathsf{coNP} \subseteq \mathsf{P/poly}$.*

**Proof:** The justification is twofold:

(1) $ForgetLit(\Sigma, L)$ is the logically strongest consequence of $\Sigma$ that is Lit-independent from $L$. Consequently, for every formula $\gamma \in PROP_{PS}$, we have $\Sigma \models \gamma$ if and only if $ForgetLit(\Sigma, L) \models \gamma$, where $L = Lit(\Sigma) \setminus Lit(\gamma)$. Because $ForgetLit(\Sigma, L)$ depends only on literals of $Lit(\Sigma) \cap Lit(\gamma)$, it is an *interpolant* of $\Sigma$ and $\gamma$, i.e., a formula $\varphi$ s.t. $\Sigma \models \varphi$ and $\varphi \models \gamma$ hold; thus we have $\Sigma \models ForgetLit(\Sigma, Lit(\Sigma) \setminus Lit(\gamma)) \models \gamma$.

(2) the existence of a propositional formula, interpolant of $\Sigma$ and $\gamma$, and of size polynomially bounded in $|\Sigma| + |\gamma|$ would imply that $\mathsf{NP} \cap \mathsf{coNP} \subseteq \mathsf{P/poly}$ (Boppana & Sipser, 1990), which is considered very unlikely in complexity theory. ◇

**Proposition 24** LIT-EQUIVALENCE *and* VAR-EQUIVALENCE *are $\Pi_2^p$-complete.*

**Proof:**

- Var-equivalence.

    - Membership: In order to check the membership to the complementary problem, guessing a clause $\gamma$ built up from $V$, and checking that ($\Sigma \models \gamma$ and $\Phi \not\models \gamma$) or ($\Sigma \not\models \gamma$ and $\Phi \models \gamma$) is sufficient. The check step can be easily accomplished in polynomial time when an $\mathsf{NP}$ oracle is available. Hence, the complementary problem belongs to $\Sigma_2^p$.

    - Hardness: Let $M$ be the mapping that associates the triple $\langle \Sigma, true, A \rangle$ to the quantified boolean formula $\forall A \exists B \Sigma$ (where $\{A, B\}$ is a partition of $Var(\Sigma)$). Clearly enough, $M$ is polytime. Moreover, we have:

$$\forall A \exists B \Sigma \text{ is valid} \quad \text{if and only if} \quad \models \exists B \,.\, \Sigma$$
$$\text{if and only if} \quad \models ForgetVar(\Sigma, B)$$
$$\text{if and only if} \quad \Sigma \equiv_A true$$

    Since the validity problem for 2-$\overline{\text{QBF}}$ formulas is $\Pi_2^p$-complete, this proves the $\Pi_2^p$-hardness of VAR-EQUIVALENCE.

- Lit-equivalence.

    - Membership: See the membership proof above, replacing "built up from $V$" by "s.t. $Lit(\gamma) \subseteq L$".

    - Hardness: Let $M$ be the mapping that associates $\langle \Sigma, \Phi, V \rangle$ to $\langle \Sigma, \Phi, L_V \rangle$. Clearly enough, $M(\langle \Sigma, \Phi, V \rangle)$ can be computed in time polynomial in $|\langle \Sigma, \Phi, V \rangle|$. We have shown that $\Sigma$ and $\Phi$ are Var-equivalent given $V$ if and only if $\Sigma$ and $\Phi$ are

Lit-equivalent given $L_V$. Hence, $M$ is a polynomial many-one reduction from VAR-EQUIVALENCE to LIT-EQUIVALENCE. Since VAR-EQUIVALENCE is $\Pi_2^p$-hard, this is also the case for LIT-EQUIVALENCE.

$\diamond$

**Proposition 25** *Let $\Sigma$ be a formula from $PROP_{PS}$ and $V$ a subset of $PS$. $\Sigma$ is influenceable from $V$ if and only if $\Sigma$ is Var-dependent on $V$.*

**Proof:** Proposition 4 from (Boutilier, 1994) states that $\Sigma$ is influenceable from $V$ if and only if there exists a prime implicant of $\Sigma$ that contains a variable from $V$, where $V$ is the set of controllable variables. Proposition 9 completes the proof. $\diamond$

**Proposition 26** *Let $\Sigma$ be a formula from $PROP_{PS}$ and $V$ a subset of $PS$. $\Sigma$ is relevant to $V$ if and only if $\Sigma$ is Var-dependent on $V$.*

**Proof:** The proof is trivial from Proposition 9. $\diamond$

**Proposition 27** *Let $\Sigma$ be a formula from $PROP_{PS}$ and $V$ a subset of $PS$. $\Sigma$ is strictly relevant to $V$ if and only if $\Sigma$ is Var-dependent on $V$ and Var-independent from $Var(\Sigma) \backslash V$.*

**Proof:** Easy from the definition of strict relevance, plus Proposition 9 ($\Sigma$ is Var-dependent from every variable occurring in a prime implicate of $\Sigma$ and Var-independent from all the remaining variables). $\diamond$

**Proposition 28**
*(1)* STRICT RELEVANCE OF A FORMULA TO A SUBJECT MATTER *(Lakemeyer, 1995) is* $\Pi_2^p$-*complete.*
*(2)* STRICT RELEVANCE OF A FORMULA TO A SUBJECT MATTER *(Lakemeyer, 1997) is* $\mathsf{BH}_2$-*complete.*

**Proof:**

- **strict relevance (Lakemeyer, 1995)**

  - Membership. Let us consider the complementary problem. Guess a clause $\gamma$, check that it does not contain any variable from $V$ (this can be achieved in time polynomial in $|\gamma| + |V|$, hence in time polynomial in $|\Sigma| + |V|$ since no prime implicate of $\Sigma$ can include a variable that does not occur in $\Sigma$). Then check that it is an implicate of $\Sigma$ (one call to an NP oracle) and check that every subclause of $\gamma$ obtained by removing from it one of its k literals is not an implicate of $\Sigma$ ($k$ calls to an NP oracle). Since only $k + 1$ calls to such an oracle are required to check that $\gamma$ is a prime implicate of $\Sigma$, the complementary problem of STRICT RELEVANCE belongs to $\Sigma_2^p$. Hence, STRICT RELEVANCE belongs to $\Pi_2^p$.

  - Hardness. Let $\{A, B\}$ be a partition of $Var(\Sigma)$ (for any formula $\Sigma$). $\forall A \exists B \Sigma$ is valid if and only if every prime implicate of $\Sigma$ that contains a variable from $A$

437

also contains a variable from $B$ if and only if every prime implicate of $\Sigma$ contains a variable from $B$ (since $Var(\Sigma) = A \cup B$) if and only if $\Sigma$ is strictly relevant to $B$.

- **strict relevance (Lakemeyer, 1997)**

    - Membership: Straightforward from Propositions 27 and 10.
    - Hardness: By exhibiting a polynomial reduction from SAT-UNSAT to STRICT RELEVANCE OF A FORMULA TO A SUBJECT MATTER. To any pair $\langle \phi, \psi \rangle$ of propositional formulas, let $rename(\psi)$ be a formula obtained from $\psi$ by renaming its variables. Obviously,
    (i) $rename(\psi)$ is satisfiable if and only if $\psi$ is.
    Now, let $new$ be a new variable, let
    $$\Sigma = \phi \wedge new \wedge \neg rename(\psi)$$
    and $V = Var(\phi) \cup \{new\}$. By Proposition 8, $\Sigma$ is Var-dependent on $V$ if and only if there is a prime implicant of $\Sigma$ mentioning a variable from $V$, i.e., if and only if $\phi \wedge \neg rename(\psi)$ is satisfiable, thus, using (i):
    (ii) $\Sigma$ is Var-dependent on $V$ if and only if both $\phi$ and $\neg \psi$ are satisfiable.
    Then, again after Proposition 8, $\Sigma$ is Var-independent from $Var(\Sigma) \setminus V = Var(rename(\psi))$ if and only if no prime implicant of $\Sigma$ mentions a variable from $Var(rename(\psi))$, i.e., if and only if $rename(\psi)$ is unsatisfiable, thus, using (i):
    (iii) $\Sigma$ is Var-independent from $Var(\Sigma) \setminus V$ if and only if $\psi$ is satisfiable.
    Thus, from Proposition 27, (ii) and (iii), we get that $\Sigma$ is strictly relevant to $V$ if and only if $\phi$ is satisfiable and $rename(\psi)$ is not.

    $\diamond$

## Appendix B: Glossary

Here is a small glossary of useful terms with the place where their definition can be found.

| | | |
|---|---|---|
| $PROP_V$ | propositional language generated by $V$ | Section 2.1 |
| $L_V$ | set of literals built up from $V$ | Section 2.1 |
| $L_V^+$ | set of positive literals built up from $V$ | Section 2.1 |
| $L_V^-$ | set of negative literals built up from $V$ | Section 2.1 |
| $Var(\Sigma)$ | set of propositional variables appearing in $\Sigma$ | Section 2.1 |
| NNF | negation normal form | Section 2.1 |
| $Lit(\Sigma)$ | set of literals occurring in the NNF of $\Sigma$ | Section 2.1 |
| $\omega_V$ | a $V$-world (instantiations of all variables of $V$) | Section 2.1 |
| $\Omega_V$ | set of all $V$-worlds | Section 2.1 |
| $\omega$ | world (full instanciation) | Section 2.1 |
| $Mod(\Sigma)$ | set of models of $\Sigma$ | Section 2.1 |
| $for(S)$ | formula such that $Mod(\Sigma) = S$ | Section 2.1 |
| $\Sigma_{x\leftarrow 0}$ | | Section 2.1 |
| $\Sigma_{x\leftarrow 1}$ | | Section 2.1 |
| $\Sigma_{l\leftarrow 1}$ | | Section 2.1 |
| $Force(\omega, l)$ | | Section 2.1 |
| $IP(\Sigma)$ | set of prime implicates of $\Sigma$ | Section 2.1 |
| $PI(\Sigma)$ | set of prime implicants of $\Sigma$ | Section 2.1 |
| $\mathsf{BH}_2$, $\mathsf{coBH}_2$ | | Section 2.2 |
| $\Delta_2^p$, $\Sigma_2^p$, $\Pi_2^p$ | | Section 2.2 |
| | syntactical Lit-dependence | Definition 1 |
| | syntactical Var-dependence | Definition 2 |
| $l \mapsto \Sigma$ | (semantical) Lit-dependence | Definition 3 |
| $DepLit(\Sigma)$ | literals such that $l \mapsto \Sigma$ | Definition 3 |
| $v \mapsto_-^+ \Sigma$ | (semantical) Var-dependence | Definition 4 |
| $DepLit(\Sigma)$ | variables such that $l \mapsto \Sigma$ | Definition 4 |
| *Lit-simplified* | | Definition 6 |
| *Var-simplified* | | Definition 6 |
| $ForgetLit(\Sigma, L)$ | literal forgetting | Definition 7 |
| $ForgetVar(\Sigma, L)$ | variable forgetting | Definition 8 |
| $\Sigma \equiv_L \Phi$ | Lit-equivalence | Definition 9 |
| $\Sigma \equiv_V \Phi$ | Var-equivalence | Definition 9 |
| | influenceability | Definition 10 |
| | relevance to a subject matter | Definition 11 |
| | strict relevance to a subject matter | Definition 12 |

# References

Amir, E., & McIlraith, S. (2000). Partition-based logical reasoning. In *Proceedings of the Seventh International Conference on Principles of Knowledge Representation and Reasoning (KR'00)*, pp. 389–400.

Biere, A., Cimatti, A., Clarke, E. M., Fujita, M., & Zhu, Y. (1999). Symbolic model checking using SAT procedures instead of BDDs. In *Proceedings of Design Automation Conference (DAC'99)*.

Boppana, R. B., & Sipser, M. (1990). The complexity of finite functions. In van Leeuwen, J. (Ed.), *Handbook of Theoretical Computer Science*, Vol. A, chap. 14. Elsevier Science Publishers (North-Holland), Amsterdam.

Boutilier, C. (1994). Toward a logic for qualitative decision theory. In *Proceedings of the Fourth International Conference on the Principles of Knowledge Representation and Reasoning (KR'94)*, pp. 75–86.

Chopra, S., & Parikh, R. (1999). An inconsistency tolerant model for belief representation and belief revision. In *Proceedings of the Sixteenth International Joint Conference on Artificial Intelligence (IJCAI'99)*, pp. 192–197.

Darwiche, A. (1997). A logical notion of conditional independence: properties and applications. *Artificial Intelligence*, *97*(1–2), 45–82.

Darwiche, A. (1998). Model-based diagnosis using structured system descriptions. *Journal of Artificial Intelligence Research*, *8*, 165–222.

Darwiche, A. (1999). Compiling knowledge into decomposable negation normal form. In *Proceedings of the Sixteenth International Joint Conference on Artificial Intelligence (IJCAI'99)*, pp. 284–289.

Darwiche, A., & Marquis, P. (1999). A perspective on knowledge compilation. In *Proceedings of the Seventeenth International Joint Conference on Artificial Intelligence (IJCAI'01)*, pp. 175–182.

Davis, M., & Putnam, H. (1960). A computing procedure for quantification theory. *Journal of the ACM*, *7*, 201–215.

Dechter, R., & Rish, I. (1994). Directional resolution; the davis-putnam procedure, revisited. In *Proceedings of the Fourth International Conference on the Principles of Knowledge Representation and Reasoning (KR'94)*, pp. 134–145.

del Val, A. (1999). A new method for consequence finding and compilation in restricted language. In *Proceedings of the Sixteenth National Conference on Artificial Intelligence (AAAI'99)*, pp. 259–264, Orlando (FL).

Doherty, P., Lukaszewicz, W., & Madalinska-Bugaj, E. (1998). The PMA and relativizing change for action update. In *Proceedings of the Sixth International Conference on Principles of Knowledge Representation and Reasoning (KR'98)*, pp. 258–269.

Doherty, P., Lukaszewicz, W., & Szalas, A. (2001). Computing strongest necessary and weakest sufficient conditions of first-order formulas. In *Proceedings of the Seventeenth International Joint Conference on Artificial Intelligence (IJCAI'01)*, pp. 145–151.

Fargier, H., Lang, J., & Marquis, P. (2000). Propositional logic and one-stage decision making. In *Proceedings of the Seventh International Conference on Principles of Knowledge Representation and Reasoning (KR'00)*, pp. 445–456.

Fariñas del Cerro, L., & Herzig, A. (1996). Belief change and dependence. In *Proceedings of the Sixth Conference on Theoretical Aspects of Reasoning about Knowledge (TARK'96)*, pp. 147–161.

Gelfond, M., & Lifschitz, V. (1993). Representing action and change by logic programs. *Journal of Logic Programming, 17*, 301–323.

Ghidini, C., & Giunchiglia, F. (2001). Local model semantics, or contextual reasoning = locality + compatibility. *Artificial Intelligence, 127*, 221–259.

Goldblatt, R. (1987). *Logics of Time and Computation*, Vol. 7 of *CSLI Lecture Notes*. Center for the Study of Language and Information, Stanford, CA.

Greiner, R., & Genesereth, M. R. (1983). What's new? a semantic definition of novelty. In *Proceedings of the Eighth International Joint Conference on Artificial Intelligence (IJCAI'83)*, pp. 450–454.

Greiner, R., & Subramanian, D. (1995). Proceedings of the AAAI fall symposium on relevance.. Technical Report FS-94-02, AAAI Press.

Hegner, S. (1987). Specification and implementation of programs for updating incomplete information databases. In *Proceedings of the Sixth ACM SIGACT SIGMOD SIGART Symposium on Principles of Database Systems (PODS'87)*, pp. 146–158.

Herzig, A. (1996). The PMA revisited. In *Proceedings of the Fifth International Conference on the Principles of Knowledge Representation and Reasoning (KR'96)*, pp. 40–50.

Herzig, A., Lang, J., Marquis, P., & Polacsek, T. (2001). Updates, actions, and planning. In *Proceedings of the Seventeenth International Joint Conference on Artificial Intelligence (IJCAI'01)*, pp. 119–124.

Herzig, A., & Rifi, O. (1998). Update operations: a review. In *Proceedings of the Thirteenth European Conference on Artificial Intelligence (ECAI'98)*, pp. 13–17.

Herzig, A., & Rifi, O. (1999). Propositional belief base update and minimal change. *Artificial Intelligence, 115*(1), 107–138.

Inoue, K. (1992). Linear resolution in consequence–finding. *Artificial Intelligence, 56*(2–3), 301–353.

Kautz, H., Kearns, M., & Selman, B. (1993). Reasoning with characteristic models. In *Proceedings of the Eleventh National Conference on Artificial Intelligence (AAAI'93)*, pp. 34–39.

Kautz, H., Kearns, M., & Selman, B. (1995). Horn approximations of empirical data. *Artificial Intelligence, 74*(1), 129–145.

Kautz, H., McAllester, D., & Selman, B. (1996). Encoding plans in propositional logic. In *Proceedings of the Fifth International Conference on the Principles of Knowledge Representation and Reasoning (KR'96)*, pp. 374–384.

Khardon, R., & Roth, D. (1996). Reasoning with models. *Artificial Intelligence*, *87*(1–2), 187–213.

Kohlas, J., Moral, S., & Haenni, R. (1999). Propositional information systems. *Journal of Logic and Computation*, *9*(5), 651–681.

Lakemeyer, G. (1995). A logical account of relevance. In *Proceedings of the Fourteenth International Joint Conference on Artificial Intelligence (IJCAI'95)*, pp. 853–859.

Lakemeyer, G. (1997). Relevance from an epistemic perspective. *Artificial Intelligence*, *97*(1–2), 137–167.

Lang, J., Liberatore, P., & Marquis, P. (2002). Conditional independence in propositional logic. *Artificial Intelligence*, *141*(1–2), 79–121.

Lang, J., & Marquis, P. (1998a). Complexity results for independence and definability in propositional logic. In *Proceedings of the Sixth International Conference on Principles of Knowledge Representation and Reasoning (KR'98)*, pp. 356–367.

Lang, J., & Marquis, P. (1998b). Two notions of dependence in propositional logic: controllability and definability. In *Proceedings of the Fifteenth National Conference on Artificial Intelligence (AAAI'98)*, pp. 268–273.

Lang, J., & Marquis, P. (2002). Resolving inconsistencies by variable forgetting. In *Proceedings of the Eighth International Conference on Principles of Knowledge Representation and Reasoning (KR'02)*, pp. 239–250.

Lang, J., Marquis, P., & Williams, M.-A. (2001). Updating epistemic states. In *Proceedings of the Fourteenth Australian Joint Conference on Artificial Intelligence (AI'01)*, pp. 297–308.

Levy, A., Fikes, R., & Sagiv, Y. (1997). Speeding up inferences using relevance reasoning: A formalism and algorithms. *Artificial Intelligence*, *97*(1-2), 83–136.

Lin, F. (2000). On strongest necessary and weakest sufficient conditions. In *Proceedings of the Seventh International Conference on Principles of Knowledge Representation and Reasoning (KR'00)*, pp. 167–175.

Lin, F., & Reiter, R. (1994). Forget it!. In *Proceedings of the AAAI Fall Symposium on Relevance*, pp. 154–159.

Marquis, P. (1994). Possible models approach via independency. In *Proceedings of the Eleventh European Conference on Artificial Intelligence (ECAI'94)*, pp. 336–340.

Marquis, P. (2000). Consequence finding algorithms. In *Handbook on Defeasible Reasoning and Uncertainty Management Systems, Volume 5: Algorithms for Uncertain and Defeasible Reasoning*, chap. 2, pp. 41–145. Kluwer Academic Publishers.

Marquis, P., & Porquet, N. (2000). Decomposing propositional knowledge bases through topics (extended abstract). In *Proceedings of the meeting on "Partial knowledge and uncertainty: independence, conditioning, inference"*.

McCarthy, J. (1986). Applications of circumscription to formalizing common-sense knowledge. *Artificial Intelligence*, *28*, 89–116.

McIlraith, S., & Amir, E. (2001). Theorem proving with structured theories. In *Proceedings of the Seventeenth International Joint Conference on Artificial Intelligence (IJCAI'01)*, pp. 624–631.

Papadimitriou, C. H. (1994). *Computational Complexity*. Addison-Wesley.

Parikh, R. (1996). *Beliefs, belief revision, and splitting languages*, pp. 266–268. Logic, Language and Computation. CSLI Publications.

Park, T. J., & Gelder, A. V. (1996). Partitioning methods for satisfiability testing on large formulas. In *Proceedings of the Thirteenth International Conference on Automated Deduction (CADE'96)*, pp. 748–762.

Przymusinski, T. C. (1989). An algorithm to compute circumscription. *Artificial Intelligence*, *38*, 49–73.

Reiter, R. (1987). A theory of diagnosis from first principles. *Artificial Intelligence*, *32*, 57–95.

Rintanen, J. (2001). Partial implicit unfolding in the Davis-Putnam procedure for quantified boolean formulae. In *Proceedings of the QBF2001 Workshop at IJCAR'01*, pp. 84–93.

Ryan, M. D. (1991). Defaults and revision in structured theories. In *Proceedings of the Sixth IEEE Symposium on Logic in Computer Science (LICS'91)*, pp. 362–373.

Ryan, M. D. (1992). *Ordered presentations of theories*. Ph.D. thesis, Imperial College, London.

Sandewall, E. (1995). *Features and Fluents*. Oxfor University Press.

Siegel, P. (1987). *Représentation et utilisation des connaissances en calcul propositionnel*. Thèse d'État, Université d'Aix–Marseille 2. (in french).

Simon, L., & del Val, A. (2001). Efficient consequence finding. In *Proceedings of the Seventeenth International Joint Conference on Artificial Intelligence (IJCAI'01)*, pp. 359–365.

Subramanian, D., Greiner, R., & Pearl, J. (1997). Artificial Intelligence Journal: Special Issue on Relevance, 97 (1-2), 1997.

Tan, S., & Pearl, J. (1994). Specification and evaluation of preferences for planning under uncertainty. In *Proceedings of the Fourth International Conference on the Principles of Knowledge Representation and Reasoning (KR'94)*.

Williams, P. F., Biere, A., Clarke, E. M., & Gupta, A. (2000). Combining Decision Diagrams and SAT Procedures for Efficient Symbolic Model Checking. In *Proceedings of the Twelfth International Conference on Computer Aided Verification (CAV'00)*.

Winslett, M. (1990). *Updating Logical Databases*. Cambridge Tracts in Theoretical Computer Science. Cambridge University Press.